# Development of FDD-ON: an Ontology for VAV HVAC System Fault Detection and Diagnostics

Yimin Chen[1]*, Brian Fricke[1], Bo Shen[1], Jamie Lian[1], Mingkan Zhang[1], James Lo[2], Yun Zhang[3], Shi Ye[4], Jiajing Huang[5], Han Hu[6], Chujie Lu[7], Rui Tang[8], George Zhuang[9]

1. Building Technologies Research and Integration Center, Oak Ridge National Laboratory, Oak Ridge, TN 37830, U.S.
2. Civil, Architectural and Environmental Engineering Department, Drexel University, 3141 Chestnut St. Philadelphia, PA 94706, U.S.
3. Lu+S Engineers. 4924 Dominion Blvd, Glen Allen, VA 23060, U.S.
4. Merck & Co. 126 E Lincoln Ave, Rahway, NJ 07065, U.S.
5. School of Data Science and Analytics, Kennesaw State University, Marietta, GA 30060, U.S.
6. Department of Mechanical Engineering, University of Arkansas, Fayetteville, AR 72701, U.S.
7. Department of Architectural Engineering and Technology, Delft University of Technology, Julianalaan 134, 2628 BL Delft, Netherlands
8. Institute for Environmental Design and Engineering, The Bartlett, University College London, London WC1H 0NN, U.K.
9. Faculty of Mathematics, University of Waterloo, Waterloo, ON N2L 3G1, Canada

* Corresponding author Email: cheny5@ornl.gov

**Abstract**

Fault detection and diagnosis (FDD) technology is essential for improving HVAC system reliability, energy efficiency, and maintenance effectiveness. However, effective deployment of FDD solutions in buildings requires structured domain knowledge that can bridge heterogeneous data sources, diverse equipment types, and varied diagnostic outputs. Limited data interpretability and interoperability within the FDD domain have led to fragmented information silos, hindering the implementation of FDD and related applications, such as the digital twin-enabled FDD frameworks and artificial intelligence (AI)-driven maintenance decision-making systems. This paper presents an FDD Ontology (FDD-ON), a modular and extensible ontology to formally represent variable air volume (VAV) HVAC system components, fault types, symptom statuses, fault impacts and associated attributes. FDD-ON integrates HVAC system FDD semantics to provide comprehensive representations of fault and symptom attributes, supported by the well-defined controlled vocabulary. Additionally, FDD-ON offers comprehensive fault, symptom, and impact libraries to capture a broad spectrum of operational abnormalities and their consequences in VAV HVAC systems. Through explicit contributing cause-fault-symptom-impact relations, FDD-ON serves as a machine-interpretable basis for querying diagnostic knowledge, mapping heterogeneous FDD outputs, and developing interoperable FDD-related applications. FDD-ON is evaluated using publicly available VAV HVAC system datasets and demonstrated through FDD development applications. Results indicate that FDD-ON provides a foundational semantic framework for advancing scalable, transparent, and interoperable FDD solutions across various applications.

## 1. Introduction

Heating, ventilation, and air conditioning (HVAC) systems are the dominant energy consumers in buildings, accounting for approximately 38% of total building energy use in the United States [1]. In addition to their substantial energy consumption and associated utility bills, the maintenance costs for ensuring system performance and improving HVAC system reliability are enormous. For example, it is estimated that the global HVAC maintenance cost was approximately $78.5 billion in 2023 and is projected to exceed $116 billion by 2030, driven by increasing demands for energy efficiency and the adoption of digital maintenance technologies [2]. Numerous faults and malfunctions occurring in HVAC systems can significantly increase energy consumption and maintenance costs [3]. For instance, studies have estimated that 15%–30% of energy in commercial buildings is wasted due to faulty, poorly maintained, degraded, or improperly controlled HVAC equipment [4–7]. Consequently, fault detection and diagnostics (FDD) technologies have become essential for improving system reliability, enhancing operational performance, reducing energy waste and supporting efficient maintenance practices. Extensive research efforts have focused on developing advanced FDD algorithms with improved fault classification accuracy, lower false alarm rates, greater scalability, and enhanced capability to address the scarcity of labelled fault data [8–12].

Despite substantial efforts on improving FDD algorithms, relatively few studies have addressed the development of knowledge models, which can enhance the development, evaluation, and integration of FDD-related solutions, such as a digital twin (DT)-based FDD deployment, semantic FDD inference frameworks, and AI-driven maintenance decision-support systems.

The absence of such frameworks presents three key challenges that impede

> (1) Lack of unified definitions of faults (or specific component malfunction), symptoms and impacts. In many FDD algorithm development, the process of FDD is to identify abnormalities or malfunctions in specific components through the manifestations of symptoms [13,14]. However, it is also reported that in FDD solutions, various symptoms and even fault impacts are classified into faults [15]. For example, in an FDD tool, an abnormal supply air flow rate in an air handling unit (AHU) is classified as a fault rather than a symptom of an underlying fault. Furthermore, an increased hot or chilled water consumption is considered as an outcome-based fault instead of a fault impact [15]. This creates ambiguity in fault definitions and hence poses challenges for establishing robust fault inference mechanisms and conducting meaningful assessments of FDD approaches.
>
> (2) Lack of sufficient illustration of attributes associated with faults and symptoms. The attributes describe fault mechanisms, symptom characteristics, as well as the cause-and-effect relations among faults, symptoms and impacts. Therefore, attribute information significantly supports FDD algorithm development and validation, subsequent maintenance activities. For example, in most HVAC FDD algorithms, symptom patterns are only represented by the deviation levels (e.g., supply air temperature in an AHU is higher/lower than the baseline or the setpoint) to indicate certain types of faults (e.g., cooling coil valve is stuck, or heating coil valve is stuck). However, other symptom patterns such as temperature slow response or oscillation, which indicate other types of faults (e.g., cooling coil water-side fouling or cooling coil valve is hunting) are often neglected.
>
> (3) The heterogeneous data outputs generated by FDD tools have limited interoperability, causing a significant challenge for the efficient interpretation and utilization of FDD results [16]. The outputs of existing FDD and system maintenance tools consist of a mix of predefined codes, abbreviations and natural language descriptions, resulting in a high volume of unstructured and inconsistently formatted information related to fault observations, symptom manifestations, diagnostic conclusions, and fault cause assessments. Such heterogeneity dramatically hinders the automated processing, integration and analysis of FDD related data, thereby restricting the ability to transform raw diagnostic outputs into actionable insights for evaluating system operational performance, supporting maintenance decision-making, and facilitating the closure of the fault management loop through effective corrective actions.

To address these challenges, researchers have increasingly turned to ontologies, which use formal, explicit conceptual models — to support semantic reasoning, knowledge integration, and explainable diagnostics. Ontologies provide a structured and machine-interpretable representation of domain knowledge, which is particularly valuable for integrating data-driven and AI technologies into HVAC system control and O&M. In the context of HVAC system FDD, ontologies can explicitly model HVAC equipment, components, sensors, operating states, faults, symptoms, and the cause-and-effect relationships among them.

As such, they provide consistent interpretation of heterogeneous data sources and facilitate the characterization of fault behaviors and occurrence mechanisms in different system configurations. Moreover, ontologies promote semantic interoperability and support explainable reasoning. This allows diagnostic approaches to combine rule-based logic, physics-informed relationships, and metadata-driven methods within a unified framework. These capabilities make ontologies as a foundational technology for developing interoperable, explainable, and scalable FDD ecosystems, which significantly support the entire fault management lifecycle, from FDD development, deployment and implementation to maintenance planning and corrective actions.

In this study, we develop an FDD ontology (FDD-ON), which creates a standardized structure for the classification of various types of faults and symptoms, as well as attributes associated with faults and symptoms. The FDD-ON unifies the representation of faults and symptoms among varied terminologies and vocabulary, into a relational ontology that permits inference and reasoning of the relationships among fault concepts and is optimized toward annotating HVAC faults.

The contributions of this work include:

(1) FDD-ON includes a semantic model to explicitly represent fault characteristics (i.e., fault descriptions and associated attributes), symptom characteristics (i.e., symptoms descriptions and associated attributes) and impact types, as well as their relations. The semantic model represents the complexity of fault mechanisms and symptom features in VAV HVAC systems. Consequently, it improves the understanding of fault mechanisms and enables the systematic dissemination of knowledge across diverse applications.

(2) FDD-ON provides a taxonomy framework, which classifies various fault attributes and symptom attributes. The developed taxonomies enhance information representations and efficient information management.

(3) FDD-ON provides definitions for fault natures and symptom statuses through controlled vocabulary items. Additionally, the naming convention for fault types, symptom statuses and impact types are structured through a quadrinomial nomenclature. This significantly enhances the interpretability and interoperability of faults and symptoms and facilitates the translation of FDD results into multiple applications.

(4) FDD-ON contains open source and evolving libraries for fault types, symptom statuses, and impacts, to efficiently retrieve fault types, as well as associated symptoms and impacts. Currently, 469 fault types, 468 symptom statuses, 447 impact types, which are reported in five major types of subsystems and equipment (i.e., chiller plant, boiler plant, AHU, VAV air terminal unit (ATU) and packaged rooftop unit (RTU)) in VAV HVAC systems, have been included in the libraries for fault types, symptom statuses, and impact types, respectively (Version: 2026B). The libraries will be continuously evolving, so that new types of faults and symptoms in VAV HVAC systems can be included and the community can incorporate the latest results for their applications.

The subsequent sections of the paper are organized as follows: Section 2 reviews existing works on the development of ontologies in the building sector. Section 3 illustrates the development of FDD-ON. Section 4 evaluates FDD-ON, as well as discusses the limitations and future work. Section 5 concludes this paper.

## 2. Literature review

To support the development of the ontology tailored to the HVAC system FDD-related applications, we reviewed existing research on the development and applications of ontologies in the building sector. The review identifies the ontology development progress and uncovers gaps in existing ontologies. Meanwhile, it helps to determine whether certain ontologies can be reused in the development of FDD-ON.

### 2.1 Ontology in the building sector

With the increasing deployment of sensors, automation systems, and diverse data sources in modern buildings, there is a need to efficiently integrate various data sources across various systems and applications in buildings. In the past, several protocols and standards, such as BACnet (Building Automation and Control Networks) [17], ASHRAE (American Society of Heating, Refrigerating and Air-Conditioning Engineers) standard 205 [18,19], 232 [20], gbXML (Green Building XML schema) [21], provide standardized way to design and document data models in various formats.

Furthermore, ontologies have become foundational in addressing heterogeneity, interoperability, and semantic understanding in building systems. A major strand of research focuses on semantic integration and data interoperability. For example, Pauwels and Fierro highlighted that metadata associated with building data streams is essential for meaningful interpretation and control in smart buildings [22]. They indicated that merging model-based metadata schemas (semantics) with data-driven monitoring and control into a functional system architecture is challenging. Pritoni et al. presented a comprehensive review on metadata schemas and ontologies used for building energy systems [23]. Their work underscores the strong needs for developing harmonization and standardization of schemas to enable interoperable, usable, and maintainable building models. A brief illustration of existing ontologies or semantic models are presented below.

The Industry Foundation Classes (IFC) provides a set of definitions for classes (entities), attributes, relations, property sets and other entities to enable semantic interoperability among Building Information Model (BIM) software platforms [24]. Although the IFC ontology provides comprehensive standard representations of equipment in the HVAC domain, it lacks a representation of FDD knowledge, which needs to describe fault mechanisms and associated symptom statuses and consequent impacts in HVAC systems.

Balaji et al. introduced Brick Schema, which provides standardized, machine-readable description of various systems in buildings [25]. In Brick Schema, a class hierarchy is defined to incorporate linked entities across physical, logical and virtual assets in building systems. Relations between entities can be represented through physical connections within the system. Several research works extended or integrated Brick Schema into various applications to enhance the data interoperability within the HVAC system-related applications. For example, Luo et al. extended the Brick Schema to represent metadata of occupants [26]. Ploennigs et al. extended the Brick Schema for automated comfort diagnosis [27]. Chia et al. reported integrating the Brick Schema into the plug load control strategies for load reduction [28]. Jia et al. developed a semantic framework for evaluating chiller plant operation using the Brick ontology [29].

Project Haystack is a widely adopted metadata framework that uses standardized tags and relations to describe building assets, sensors, points, and controls [30]. It is particularly useful for addressing the inconsistent point naming conventions from different vendor-specific building automation systems (BAS) and building energy management systems (BEMS) and supporting interoperable building analytics across vendor-specific systems. Nevertheless, Project Haystack is primarily designed as a general building metadata and tagging framework rather than an HVAC FDD domain ontology. Its core vocabularies do not provide information of faults in HVAC systems. Therefore, additional FDD-specific semantic structures are needed to represent diagnostic knowledge in a reusable and machine-interpretable form.

Additionally, other researchers developed various ontologies for different applications in the building system operation and control. For example, Li et al. developed Energy Flexibility Ontology (EFOnt), which extends existing terminologies, ontologies, and schemas for building energy flexibility applications [31]. Schneider et al. developed a control ontology (CTRLont) to define the domain of control logic in BAS [32]. Qiu et al. developed Human Comfort in Building ontology to capture human experiences related to various factors in the indoor environmental quality domain so that HVAC system settings can be adjusted [33]. Rasmussen et al. developed Building Topology Ontology (BOT), which provides a high-level description of the topology of buildings including stories and spaces, the building elements, and web-friendly 3D models [34]. Tomašević et al. developed a facility ontology to model all static knowledge, including technical vendor data, proprietary data types, and communication protocols in the facility management activities of airports [35].

Overall, literature highlights the benefits of ontologies used in buildings, i.e., ontologies promote semantic interoperability across solution vendors and systems. Meanwhile, ontologies provide a foundation for advanced analytics and reasoning, and facilitate the integration of complex datasets, which is essential for scalability and robust automation in smart buildings.

## 2.2 Ontology for HVAC FDD

In the context of FDD, semantic ontologies can support advanced reasoning and inference-based detection [36]. However, few studies have been conducted to develop ontologies or semantic models for HVAC system FDD. Existing ontologies and semantic models aim to enable consistent data integration and automated reasoning for HVAC system FDD tasks. For example, Chen et al. developed fault taxonomy for three types of HVAC equipment, i.e., air handling unit, terminal unit and packaged roof top unit [37,38]. In the taxonomy, the authors not only developed a standardized and structured naming system to represent various fault types in various locations of a component in equipment but also the defined control vocabulary to describe fault natures in HVAC systems. However, the authors classified fault symptoms into fault types (i.e., behavior-based fault type and outcome-based fault type), leading to difficulties in the reference process and fault types.

Hwang et al. extended Brick Schema to a Fault-Symptom Brick metadata schema, namely FSBrick [39,40]. In the schema, FSBrick provides a semantic framework that links HVAC faults, symptoms, system components, and measurable points, enabling more systematic and explainable FDD results. However, the schema did neither fully capture fault mechanisms, such as contributing causes and occurrence stages, nor fault symptom characteristics, such as symptom directions and symptom patterns.

Blechmann et al. developed fault ontology for AHUs to depict part of fault characteristics such as fault occurrence locations and fault nature [41]. The effectiveness of the developed approach was demonstrated using historic operation data of a real AHU in the building. However, the authors did not completely illustrate the structure of the ontology and provide a fault type library.

Gourabpasi et al. developed an ontology for automated FDD of HVAC, namely AFDDOnto [42]. The developed AFDDOnto has two functionalities: 1) integrating BIM with BAS concepts in the form of a knowledge model to assist AFDD model development; and 2) obtaining analytic results from AFDD models to update BIM-based model for the application of DT of the facility. Although the study developed a semantic model using concepts from BIM and BAS, it lacked a detailed description of fault natures, making it hard to implement reference during the diagnosing process in FDD.

Ploennigs et al. employed semantic graphs to create the diagnostic model from data points in HVAC systems and identify potential cause-and-effect relations [43]. Although the study differentiated the fault type and associated symptoms, it did not further illustrate the inherent mechanisms of each fault type and various symptoms. Additionally, it did not propose a data format that can facilitate streamlining the data analytics in FDD solutions and CMMSs.

Mallak et al. developed a system ontology which consists of four entities or classes i.e., office building, sensory and control, symptoms and failures classes [44]. The study was demonstrated to be effective in building a diagnostic direct graphic for fault diagnosis reference. However, the study did not provide a unified and structured data format to represent the developed ontology, making it difficult for CMMSs to automatically process FDD results.

Li et al. developed a domain ontology, which included semantic rules based on the knowledge graph for building energy system fault detection [45]. Through the ontology model, fault action mechanisms can be captured based on the inference chains of the rules. While the study defined five top classes, i.e., Equipment, Sensor, Datapoint, Location and Fault to illustrate the fault detection process, it lacked detailed descriptions on fault nature and other types of measurements such as control signals and meter data to represent fault symptoms and impacts.

## 2.3 Limitations of existing work

From the above studies, we identified three major limitations of existing ontologies. They are:

(1) Existing ontologies or semantic models have insufficient descriptions of mechanisms and attributes of faults and symptoms in HVAC systems.

(2) Non ontologies provide an open-source fault library and symptom library that cover reported fault types in HVAC systems. This prevents ontology reusability for other application developments.

(3) Existing ontologies do not consider various applications, such as the development of maintenance decision-making systems and fault tolerant control solutions, other than FDD approach development; and hence, have weak inference capabilities to support multiple applications.

## 3. Methodology

In this Section, we illustrate the method framework for developing FDD-ON as illustrated in Figure 1. There have been many approaches for developing ontology for various applications in the past years, and they were reviewed and compared in [46–48]. In this study, we adopted Gruninger & Fox's methodology [49], which transforms informal scenarios into formal languages. Specifically, we propose a framework that includes five major phases including: 1) information and requirement analysis, 2) development ontology, 3) development of fault libraries, symptom libraries, and impact libraries, 3) ontology evaluation, and 5) limitation discussion and evolution. It is noted that the development of FDD-ON is an iterative process, i.e., the evaluation result of FDD-ON should feedback to step #1 to update, extend and enhance the ontology. We illustrate this methodology in the following subsections in Section 3 and Section 4, respectively.

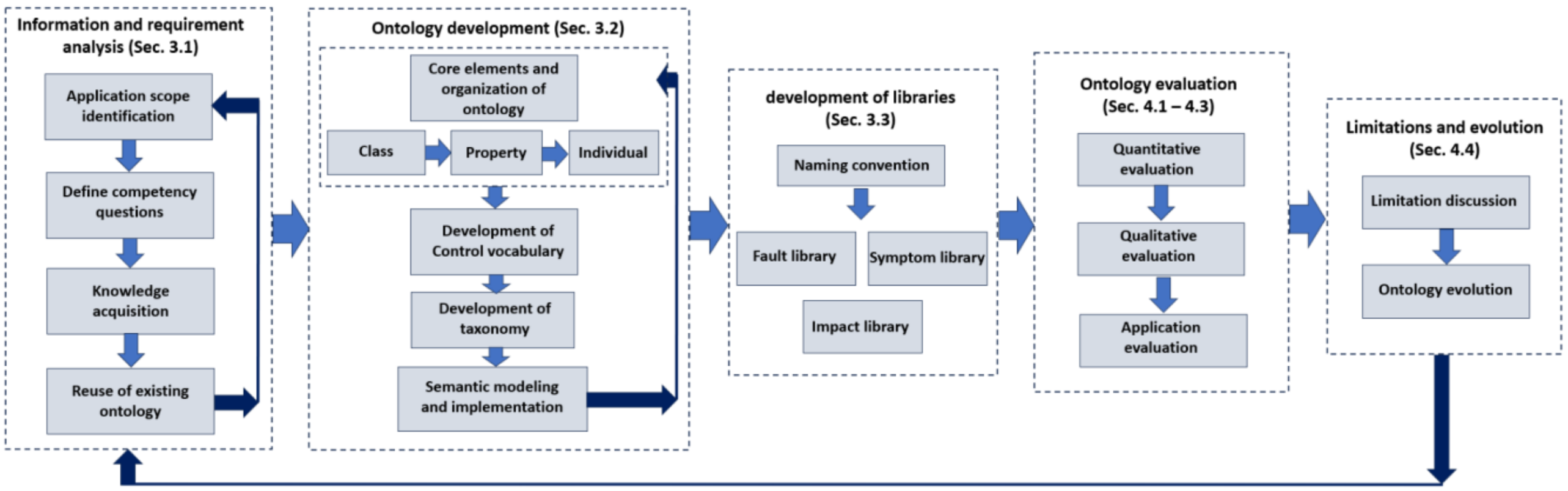

**Figure 1**. Methodology framework for developing FDD-ON.

### 3.1 Information and requirement analysis

#### 3.1.1 Application scope identification

In this study, we identify the application scope by first constructing the knowledge structure for various applications in the FDD domain. There are six fundamental directions that the FDD area includes.

**(1) Fault modeling:** Fault modeling is to construct an accurate representation of occurrence mechanisms (including the fault occurrence location, and fault behaviors) of hardware faults or software faults, as well as the associated fault symptoms and impacts [50,51]. To improve fault modeling outcomes, it is needed to clearly understand fault behaviors (e.g., fault occurrence locations) and their correlations with fault symptoms (e.g., how fault severity level may affect fault symptom states and magnitudes).

**(2) Fault effect and impact analysis:** The fault behavior and impact analysis investigates how HVAC equipment and systems operate under faulty operations and consequently uncovers the characteristics and effects of various types of faults in the equipment and systems [3,52]. To better evaluate fault effects and impacts, it is essential to clearly scope the attributes of fault symptoms and impacts. For example, in an HVAC system, for the same type of a fault, effects and impacts vary significantly under various operational conditions. Therefore, the uncertainty of fault symptom magnitudes and impact levels should be illustrated during fault effect and impact analysis.

**(3) FDD development:** The development of FDD solution often includes two phases as: i) the abnormality detection process is to capture the operational abnormalities of a system or equipment; and ii) the fault diagnosis is to identify the fault location, distinguish the fault type, and estimate fault severity levels [53–55]. To enhance the development of FDD solutions, especially for the inference-driven fault diagnostics, it is needed to establish a robust understanding of the relations between faults and associated symptoms.

**(4) Fault tolerant control implementation:** A fault tolerant control is capable of tolerating faults in the HVAC system to improve the reliability and availability while ensuring a desirable operation performance [56–59]. When a fault occurs, the HVAC control system can isolate fault source, eliminate the fault, and is robust to mitigate fault negative impacts [60]. To develop fault tolerant control strategies, it is necessary to completely understand fault occurrence mechanisms (e.g., fault occurrence location, fault severity level and fault behaviors), as well as fault symptoms and impacts.

**(5) Maintenance practice:** After faults are detected and diagnosed, the facility team or technicians should address the issue, fix the faults, or plan other maintenance activities to ensure the operational performance of the HVAC system. The maintenance algorithms need to effectively support various maintenance strategies (e.g., condition-based maintenance, preventive maintenance and predictive maintenance) based on various factors such as fault behaviors, fault impact magnitude, system operation schedules and facility team arrangement [61,62]. To support HVAC system maintenance decision-making and actual maintenance work, it is required to precisely diagnose fault types, accurately estimate fault impacts and infer fault contributing causes.

**(6) Fault diagnosis analysis:** Fault diagnosis analysis investigates fault occurrence likelihood, fault occurrence rate, fault contributing causes in HVAC systems. It provides insight into fault occurrence distributions and patterns in different types of equipment or systems under various operational conditions. Hence, the analysis results support multiple applications such as the development of FDD approaches, equipment improvement and maintenance practices [7,63]. To perform fault occurrence analytics, it is required to obtain complete descriptive information from fault records including fault types, fault behaviors, fault impacts, and how the fault is detected and diagnosed.

### 3.1.2 Competency questions

Competency questions (CQs) include a set of questions stated and replied in natural language to support the development, evaluate and maintain the ontology [64]. CQs play a critical role in the development of ontology because they represent the ontology requirements [65]. After identifying the application scope, we designed seven major categories of CQs, which are used to specify ontology functional requirements and guide the development of FDD-ON as given in Table 1. Under each category of CQs, specific CQs were designed but not listed here. For example, under CQ3, a specific CQ of 'How to classify fault contributing causes was designed to enable the developers to specify the purpose and function of classifying fault contributing causes.

**Table** 1. List of CQs.

| CQ category | CQ description |
|---|---|
| CQ1 | How to define fault types and associated symptom statuses in control systems, specifically for HVAC systems in residential buildings and commercial buildings, for the purpose of sharing knowledge? |
| CQ2 | What existing ontologies can be reused to develop the FDD ontology? |
| CQ3 | What attributes should be included to accurately describe faults found in HVAC systems? |
| CQ4 | What attributes should be included to accurately describe symptoms observed in HVAC systems? |
| CQ5 | What taxonomic hierarchy can be used to accurately classify those attributes associated with fault types, symptom statuses, and impact types? |
| CQ6 | How can the interpretability and interoperability of the fault data be enhanced to facilitate multiple applications? |
| CQ7 | How can the HVAC system fault related data be structured and organized to address the issue of data silos? |

### 3.1.3 Knowledge acquisition

Ontology development requires a thorough understanding of the target domain and its associated terminology through the collection and analysis of existing documents, databases, standards, expert knowledge, and relevant ontologies [66]. In this study, the knowledge acquisition process for developing FDD-ON was informed by three primary sources. First, we conducted extensive research work in several directions in the HVAC FDD area, including fault impact analysis [3,67], fault model development and fault data curation [51,68,69], physics-based and data-driven FDD approach development and deployment [8,70–74], fault tolerant control [75,76], fault prevalence and occurrence characteristics [7,63], facility maintenance [16], and data tagging and relation inference technologies used in FDD [38,77]; Secondly, we examined several commercialized FDD solutions to obtain an in-depth understanding of functionalities of the software tools [38,78]; Lastly, we incorporated insights obtained from facility operators and maintenance personnel to capture first-hand operational experiences and identify practical challenges encountered in day-to-day O&M activities.

These knowledge sources ensured that FDD-ON reflects both theoretical advancements and real-world requirements, thereby enhancing its applicability to practical HVAC FDD and maintenance scenarios.

### 3.1.4 Reuse of existing ontologies

In this study, we reuse some approaches and data-modeling in terms of HVAC system hierarchy and data point tagging in Project Haystack [30] and Brick Schema [25]. For example, the classification in the temperature sensor class in the Brick model was used. However, it should be noted the data point naming structure in the fault naming system was adjusted to more accurately reflect the fault and symptom mechanisms including locations, measurement types, behaviors and states. Additionally, we reused and adjusted the controlled vocabulary library and data structure that was proposed in HVAC system fault taxonomy [37]. This is because the controlled vocabulary library and the fault naming data structure developed in fault taxonomy basically capture the fault characteristics and physical configurations in HVAC systems

## 3.2 Development of ontology

### 3.2.1 Fundamentals

HVAC system FDD relies on diverse sources of information, such as operational time-series data and metadata, to support complex reasoning and cause-and-effect analyses. Based on the concept of causal chains [79,80], we developed a four-level semantic model, which characterizes the FDD inference and analytic processes, as illustrated in Figure 2. The model structures FDD knowledge and reasoning into four interconnected levels: 1) symptom detection, 2) fault diagnosis, 3) contributing cause identification, and 4) fault impact analysis.

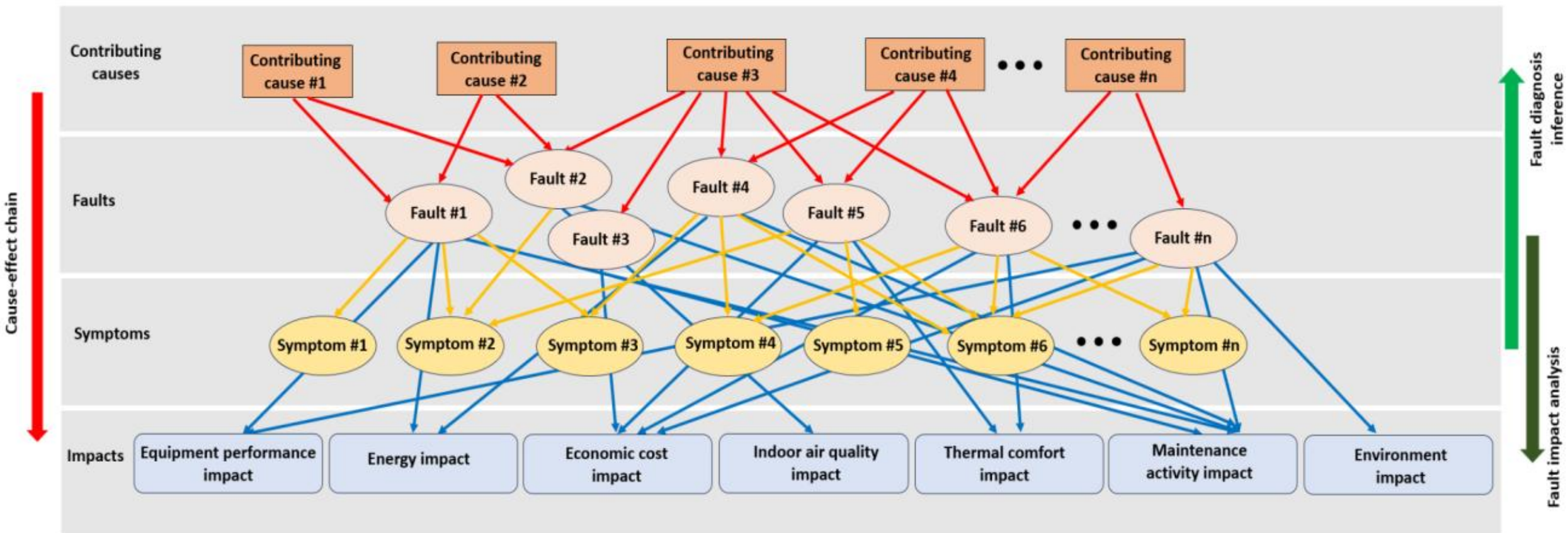

**Figure** 2. Four-level semantic model for FDD inference and analytics.

(1) **Symptom detection**: During system operation, symptoms are manifested as abnormal patterns in specific measurements identified by comparing the real-time interval data with predefined thresholds or expected baselines (e.g., the supply temperature is higher than expected). Accordingly, symptom detection aims to identify and flag these abnormal manifestations when thresholds are overpassed. Successfully symptom detection relies on accurately establishing performance baselines and appropriately defining detection thresholds [67,81]. Otherwise, missing detection and false alarms may occur. However, symptom manifestation in HVAC systems is often uncertain because it can be influenced by many factors such as fault severity levels, varying indoor and outdoor environmental conditions, and changes in operational sequences [82]. For instance, an AHU cooling valve stuck at a low position may not produce strong fault symptoms during the transition season, when cooling demand is less [54].

(2) **Fault diagnosis**: In FDD-ON, a fault refers to undesirable operational condition deviation from a defined operational goal, caused by non-functional component (e.g., a leaking valve), improper software settings (e.g., incorrect proportional-integral-derivative (PID) parameter setting), or ill-designed operational processes conditions (e.g., an improper sized pump). Consequently, the objective of fault diagnosis is to distinguish the abnormal operating condition in terms of both its nature and location in the system. One of the major challenges in fault diagnosis is that different faults may generate similar or overlapped symptom patterns [83,84]. For example, in an AHU, a negative bias fault in the supply air static pressure sensor and outdoor air damper stuck at a higher open position can produce similar abnormalities in supply air fan speed, fan power, and cooling coil valve position. Such symptom overlaps complicate the accurate identification of a fault.

(3) **Contributing cause identification:** Contributing causes represent the underlying conditions, actions, or environmental circumstances that lead to an occurrence of a fault. The process of uncovering fault contributing causes is to reveal the "why" or "how" behind a specific fault. The identification of contributing causes of a fault is critical to effectively carry out maintenance activities (e.g., estimate maintenance labor efforts and financial costs). For the same fault, the maintenance activities can be significantly different. For instance, for an AHU damper stuck fault, if the contributing cause of this fault is disconnected communication wires of the damper controller, then the maintenance work is to reconnect the communication wires, and there is very little maintenance financial cost associated with this maintenance activity. In contrast, if the fault is caused by a malfunctional damper actuator, component replacement may be necessary, leading to substantially higher maintenance costs and increased labor efforts.

Despite its importance, very few efforts have been made to investigate fault contributing causes in HVAC FDD. This is partly because identifying contributing causes often require complex causal reasoning and domain expertise that cannot be readily automated using existing FDD tools. In addition, records documenting fault contributing causes are rarely available in practice, making it difficult to establish systematic knowledge and standardized representations.

(4) **Fault impact analysis:** Faults occurring in specific components can propagate through HVAC systems and produce broader impacts on operational performance, energy consumption, occupant comfort, maintenance activities, and financial outcomes. Following fault diagnosis, fault impact analysis aims to quantify these consequences using various performance metrics [3]. Although the distinction between symptoms and impacts is not always explicit, symptoms generally represent observable indicators of abnormal operation, whereas impacts describe the resulting performance degradation or associated consequences. For instance, abrupt increases in power consumption or immediate reductions in cooling capacity may serve both as observable symptoms and as indicators of fault impacts. Nevertheless, fault impacts typically represent broader effects, including energy penalties, increased maintenance requirements, reduced equipment reliability, and occupant discomfort. When faults cannot be fixed rapidly, mitigating their impacts becomes essential and may require the implementation of sophisticated fault-tolerant control strategies.

In summary, the underlying cause-and-effect chain can therefore be summarized as follows: **contributing causes lead to faults, faults manifest through symptoms, and faults subsequently result in impacts**. In contrast, the FDD inference process proceeds in the reverse direction, beginning with symptom detection, followed by fault diagnosis, contributing-cause identification, and fault impact analysis. This reasoning framework provides the conceptual basis for the semantic model developed in the following sections.

### 3.2.2 Semantic modeling

There is a set of technologies supporting the creation of semantic models. In this study, we used RDFS (Resource Description Framework Schema) and OWL (Web Ontology Language), which are based on the World Wide Web Consortium's (W3C) standards, to develop the semantic model for data interchange on the web that allows for modeling graphs of resources over the web [85,86]. RDFS is the upgrade of Resource Description Framework (RDF), which describes resources using undirected and network graphs in the form of Triples, i.e.,three-component '*Subject-Predicate-Object*' statements which are also represented as tuples ***t***: <***s***, ***p***, **o**>.

To implement RDFS, knowledge about faults, symptoms and their relations in HVAC systems is translated into linguistic components, which are represented by logical predicates, variables, and quantifiers, to ensure the conceptualization is sound, consistent, and less ambiguous. For example, the fault and the associated fault contributing cause can be expressed in a Triple form as:

| ***<Fault>*** | ***<hasContributingCause>*** | ***<ContributingCause>*** |
|---|---|---|
| (subject) | (predicate) | (object) |

Furthermore, the fault and the associated symptoms can be expressed in a Triple form as:

| ***<Fault>*** | ***<hasSymptom>*** | ***<Symptom>*** |
|---|---|---|
| (subject) | (predicate) | (object) |

In addition to RDFS, OWL is used to describe complex relations among the fault attribute class and the symptom attribute class. OWL is built on the top of RDFS and provides more comprehensive and complicated characteristics for creating ontologies, which require much more expressiveness than RDF and RDFS. In addition, OWL offers computational logics, namely Description Logics (DL), which facilitate advanced reasoning and data integration.

The core components of OWL include **Classes**, **Properties**, and **Individuals**. Figure 3 shows the structure examples of the OWL ontology for representing a fault type and a symptom status, respectively.

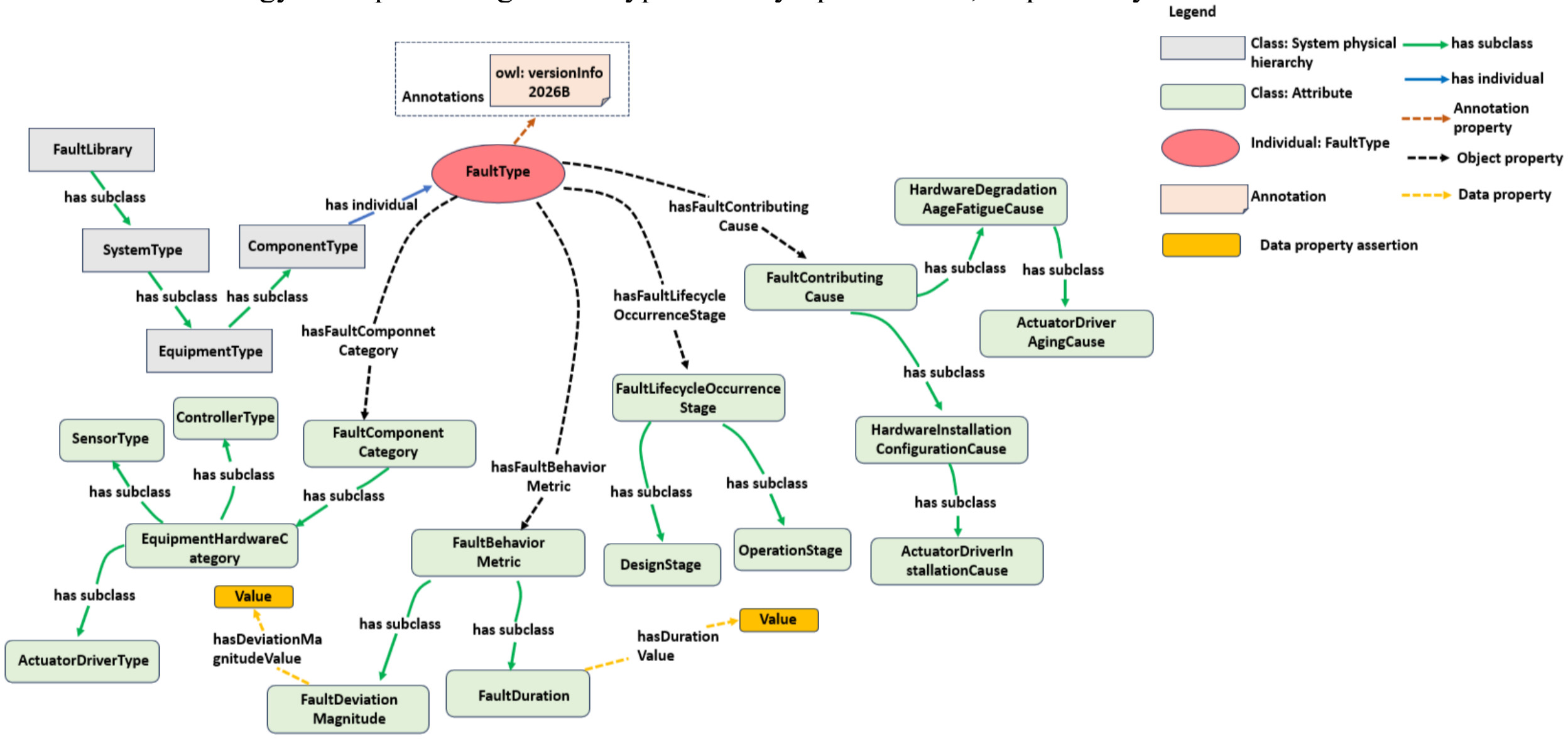

(a) A fault type and associated attributes

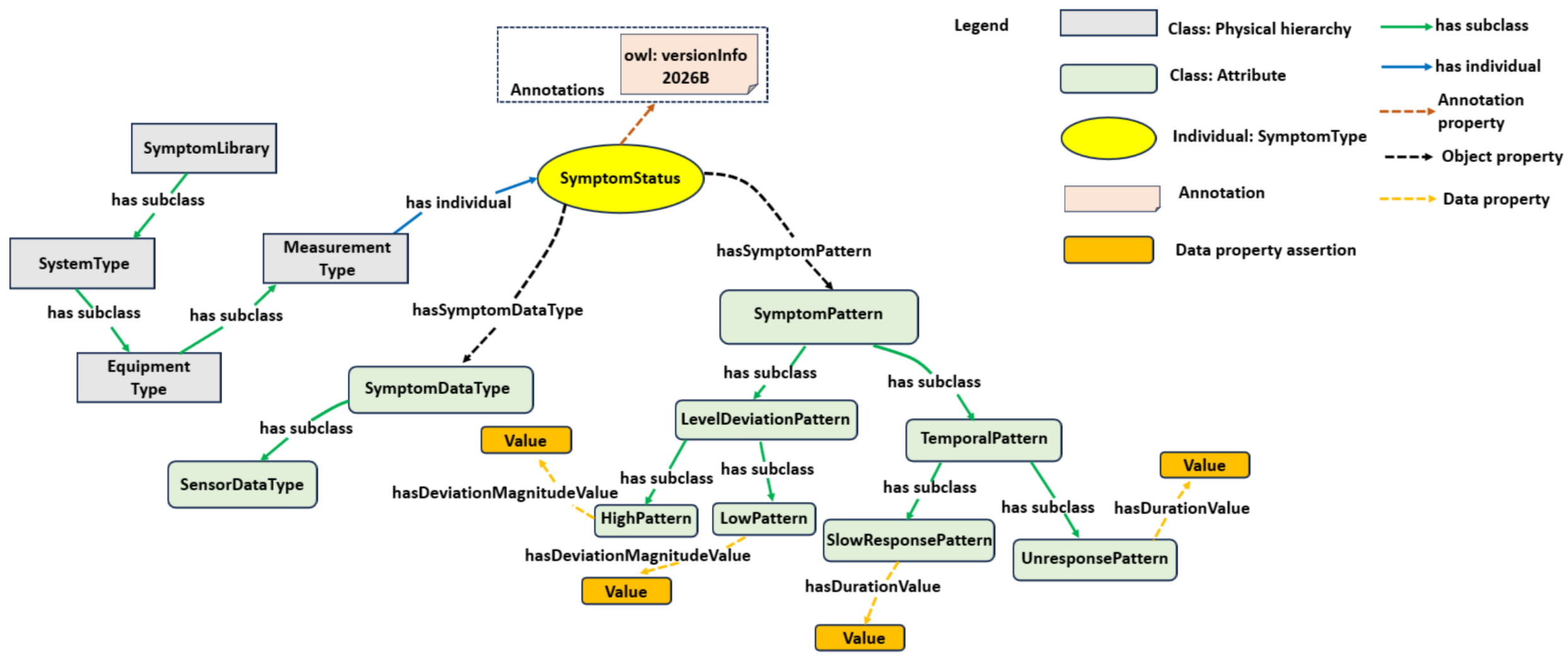

(b) A symptom status and associated attributes

**Figure** 3. The structure of FDD-ON for representing a fault type and a symptom status.

(1) **Individuals:** In OWL, individuals describe specific objects that are included in the ontology as indicated by the oval symptom in Figure 3. For example, in FDD-ON, individuals can have a specific fault type (e.g., AHU cooling coil valve leakage), or a specific symptom status (e.g., high AHU supply air temperature). In FDD-ON, fault type individuals and symptom statuses are strictly defined using the developed control vocabulary to eliminate ambiguity and enhance data interpretability as illustrated in Section 3.2.2.

(2) **Classes:** Classes categorize individuals with similar characteristics and organize them using formal descriptions in a specific domain as indicated by the rectangular block in Figure 3.

In FDD-ON, classes are built to represent two types of objects, such as i) the classes that categorize equipment, components and associated fault types and symptom statuses in the VAV HVAC systems; and ii) the classes that categorize fault attributes, symptom attributes and impact types. The classes contain single-level or multiple-level structures, which enable more granular classifications of system physical hierarchies, as well as certain attributes associated with faults and symptoms. However, it is noted that in the development of ontologies, the granularity of classes is not permanent but is determined by a potential application of the ontology.

The classes, which group similar types of equipment and components of HVAC systems, follow the physical hierarchy of the system and have three levels (i.e., sub classes) including: i) system types, ii) equipment types, and iii) component types or measurement types. For example, various types of sensors such as a discharge air temperature sensor, a relative humidity sensor, and an air flow rate sensor are categorized into the component SubClass of 'Sensor'.

In addition to the classification of physical systems, another set of classes represents similar attributes associated with a fault type or a symptom status. For instance, symptoms have an attribute set of classes, which categorize various characteristics (e.g., data types) associated with symptom statuses.

The classes are systematically structured to form the taxonomies for fault types, symptom statuses and impact types as illustrated in detail in Sections 3.2.4 to 3.2.6.

(3) **Properties: FDD-ON employs** OWL properties to link classes and individuals in the form of binary relations, which reflect the '***Predicate***' statement in RDFS. As shown in Figure 3, properties connect between classes and individuals or among individuals using arrow symbols. In FDD-ON, we employed three types of properties (the *Object Property*, the *Data Property*, and *Annotation Property*) as listed in Table 2.

**Table** 2. List of properties.

| Property name | Property type | Characteristics | SubProperty of | Property domain | Property range |
|---|---|---|---|---|---|
| hasFaultAttribute | Object | NA | NA | NA | NA |
| hasFaultBehaviorMetric | Object | NA | hasFaultAttribute | FaultLibrary | FaultBehaviorMetric |
| hasFaultComponentCategory | Object | NA | hasFaultAttribute | FaultLibrary | FaultComponnetCategory |
| hasFaultLifecycleOccurrenceStage | Object | NA | hasFaultAttribute | FaultLibrary | FaultLifecycleOccurrenceStage |
| hasFaultContributingCause | Object | NA | hasFaultAttribute | FaultLibrary | FaultContributingCause |
| hasSymptomAttribute | Object | NA | NA | NA | NA |
| hasSymptomPattern | Object | NA | hasSymptomAttribute | SymptomLibrary | SymptomPattern |
| hasSymptomDataType | Object | NA | hasSymptomAttribute | SymptomLibrary | SymptomDataType |
| hasSymptom/isSymptomOf | Object | Inverse | NA | FaultLibrary | SymptomLibrary |
| hasImpactType | Object | NA | NA | NA | NA |
| hasImpact/isImpactOf | Object | Inverse | hasImpactType | FaultLibrary | ImpactLibrary |
| hasDeviationMagnitudeValue | Data | xsd:integer | NA | NA | NA |
| hasDurationValue | Data | xsd:integer | NA | NA | NA |
| hasRecurrenceIntervalValue | Data | xsd:integer | NA | NA | NA |
| hasFrequencyValue | Data | xsd:decimal | NA | NA | NA |
| hasProbabilityValue | Data | xsd:decimal | NA | NA | NA |
| hasStrengthennessValue | Data | xsd:string | NA | NA | NA |
| rdfs:label | Annotation | String | NA | NA | NA |
| rdfs:comment | Annotation | String | NA | NA | NA |
| owl:versionInfo | Annotation | xsd:integer | NA | NA | NA |

The *Object Property* shows a relationship among various individuals. For example, the statement of 'a fault has an associated symptom' is defined by a property 'hasSymptom'. This property has an inverse property 'isSymptomOf', which represents that a specific symptom is linked to a fault.

The *Data Property* represents a relationship between an individual and associated data type. For instance, a quantitative symptom magnitude associated with a symptom is built through a data property, namely '*hasDeviationMagnitudeValue*'. It is noted that the data values associated with individuals should be customized in real practice to reflect real life operation of a system.

The *Annotation Property* illustrates meta-data such as the label and the ID of a specific fault type. The annotation properties have

### 3.2.3 Controlled vocabulary

In the HVAC FDD area, many technical terms and descriptions are used by different stakeholders to represent the same type of faults or similar symptoms [37]. This increases ambiguity and can significantly hinder the interpretability of fault information. Therefore, it is necessary to develop a set of controlled vocabulary to provide a more curated, standardized set of terms to define both the fault mode and symptom status, as well as the relationships between faults and associated symptoms in a consistent manner.

In FDD-ON, we reused and extended the controlled vocabulary list [37], which was developed to standardize the fault description. The controlled vocabulary can represent fault types and symptom statuses identified in the HVAC operations. This reduces the ambiguity and consequently, enhances the preciseness of the semantic meanings of fault types and symptom statuses, the interpretability of reporting faults and symptoms, as well as the interoperability of data across different systems. The controlled vocabularies will be encoded into a solid naming structure as illustrated in Section 3.3.

**(1) Controlled vocabularies of fault types**

Each fault vocabulary describes a specific fault type, which has unique occurrence mechanisms in terms of fault components, occurrence locations, and behaviors. In the study, we summarized 20 controlled vocabulary items, which qualitatively illustrate fault behaviors in the FDD-ON, as listed in Table 3.

**Table** 3. Controlled vocabulary list for describing fault modes.

| No. | Fault type | Definition | Example component(s) in which a fault may occur |
|---|---|---|---|
| 1 | Bias | Sensor reading value consistently deviates from the true value, creating a persistent measurement error. | Sensor |
| 2 | Drift | Sensor reading value gradually deviates from the true value over time. | Sensor |
| 3 | Freeze | Sensor reading remains a constant value even if the true value changes. | Sensor |
| 4 | Air-side fouling | Fouling is the buildup of sediments and debris on the surface area of a coil. | Coil |
| 5 | Water-side fouling | The accumulation of unwanted deposits on the internal (water-contacting) surfaces of coils. | Coil |
| 6 | Leakage | Fluid leaks from a high-pressure region to a low-pressure region. | Coil, valve, duct, damper |
| 7 | Stuck | The actuator fails to respond to changing control signals and remains stuck at a fixed point. | Valve, damper |
| 8 | Hunting | The actuator continuously modulates back and forth, over-correcting its position while attempting to reach a stable setpoint. | Valve, damper |
| 9 | Blockage | The component is physically blocked, causing a significant increase of resistance to airflow. | Filter, vent |
| 10 | Looseness | The physical part of a system is not securely fastened, properly tensioned, or correctly seated. | Fan belt |
| 11 | Failure | The component stops work. | Compressor |
| 12 | Setting error | The parameters, control sequence, or logic in the software are wrongly set. | Schedule |

| 13 | Location error | The component is installed in the wrong place. | Discharge air temperature sensor |
|---|---|---|---|
| 14 | Overcharge | The fluid is overcharged. | Refrigerant |
| 15 | Undercharge | The fluid is undercharged. | Refrigerant |
| 16 | Restriction | The flow in the component (coil, tube, or pipe) is restricted, reducing the fluid flow rate. | Refrigerant circuit |
| 17 | Loss | In communication, data packages are lost. | Communication |
| 18 | Oversize | The equipment or components are oversized. | Pump |
| 19 | Undersize | The equipment or components are undersized. | Pump |
| 20 | Offline | The controller stops work. | Controller |

**(2) Controlled vocabularies of symptom statuses**

The symptom vocabularies should accurately represent how symptoms manifest and usually indicate distinct fault types. We systematically investigated various fault symptoms and impacts reported in literature [3] and FDD software solutions [38], and then summarized 11 vocabulary items, which qualitatively describe specific fault symptom statuses in HVAC systems. The quantitative values corresponding to each status can be included under each application case. Table 4 lists the controlled vocabulary items.

**Table** 4. Controlled vocabulary list for describing symptom statuses.

| No. | Symptom status | Definition | Example measures in which the symptom may manifest |
|---|---|---|---|
| 1 | High | The measured value is higher than the baseline value or the reference value. | Supply air temperature |
| 2 | Low | The measured value is lower than the baseline value or the reference value. | Supply air temperature |
| 3 | Slow response | The measured value has excessive settling time. | Chilled water supply temperature |
| 4 | Unresponse | The measured value keeps constant when it should change. | Damper position signal |
| 5 | Oscillation | The measured value has cyclical variation. | Discharge air damper control signal |
| 6 | Cycling | The measured equipment status or equipment control command signal is switched between an 'on' (logic high or 100% of the normal value) status and an 'off' status (logic low or 0% of the normal value) frequently or irregularly. | Damper position signal |
| 7 | Spike | The measured value contains a transient sharp peak. | Pump control signal |
| 8 | Step change | The measured value appears as a sudden and discrete shift from one steady level to another level. | Damper control signal |
| 9 | Intermittent | The measured value appears and disappears irregularly. | Pump control signal |
| 10 | Mismatch | The measured values are not matched. | Damper position and control signal |

| 11 | Simultaneous Occurrence | The signals can be measured at the same time, but logically they should not be measured at the same time. | Cooling valve position signal and heating valve position signal |
|---|---|---|---|

### 3.2.4 Taxonomy for fault attributes

The fault attributes contain all the information that further describe fault characteristics and will be enriched by various applications. The taxonomies of fault attributes classify various fault attributes in an organized way. Currently, fault attributes are categorized into four classes as: 1) fault behavior metric; 2) component category; 3) contributing cause; and 4) lifecycle occurrence stage, as illustrated in Figure 3.

**(1) Fault behavior metric**

The class of fault behavior metrics defines how a fault behaves dynamically on a multi-dimensional temporal and quantitative scale, i.e., this subclass represents characteristics of the fault intensity, occurrence and evolution. Hence, this class enables an AI-driven maintenance system to quantify fault intensity, and to distinguish between transient fault and long-standing, degrading mechanical issues. It also enables the development of FDD inference approaches by providing information such as the fault occurrence likelihood. The class of fault behaviors includes four second level subclasses of behavior descriptions, as listed in Table 5.

**Table** 5. List of fault behavior metrics.

| Subclass | Annotation | Example of mapped faults |
|---|---|---|
| Fault deviation magnitude | The physical severity or intensity of the fault | Cooling coil valve damper stuck at 65% position |
| Fault duration | The length of time that a fault condition persists in a system | Discharge air damper stuck for 170 hours |
| Fault occurrence frequency | The frequency that a specific fault appears within a given time period | Supply air temperature sensor bias fault occurrence at 0.01% probability |
| Fault recurrence interval | The time interval between two consecutive occurrences of the same fault | Outdoor air damper stuck fault recurrence interval is 15,000 hours |

**(2) Component category**

The subclass of component categories categorizes functional components, where a fault may occur, into organized groups. In FDD-ON, the component category has three sub classes, including: 1) equipment software, 2) equipment hardware, and 3) network and communication. Under each sub class, the second-level subclasses that represent specific component types are defined, as listed in Table 6.

**Table** 6. List of component categories.

| Subclass (component category) | Second level subclass (component type) | Example components |
|---|---|---|
| Equipment software | Control sequence and logic | VAV air terminal schedule setting |
| | Control parameter | AHU supply air fan PID parameter |
| Equipment hardware | Sensor | AHU supply air temperature sensor |
| | Actuator and driver | VAV discharge air damper |
| | Coil | AHU cooling coil |
| | Duct | AHH supply air duct |
| | Pipe | Chiller plant chilled water pipe |
| | Vent | RTU return air vent |
| | Power supply wire and connector | RTU condenser fan power supply wire |
| | Ground wire and shield | Ground wire |
| | Controller | AHU controller |
| | Fluid | Refrigerant |
| | Fin | Heat exchanger |
| | Filter | AHU supply air filter |
| | Controller | DDC controller |
| | Electrical component | Contactor |
| Network and communication | Network interface | Router |
| | Communication wire and connector | Communication wire |
| | Communication parameter | Network address |
| | Communication shield | Wire shield |

**(3) Contributing cause**

The class of fault contributing causes categorizes the underlying factors or events that create the environment for a specific fault to occur. For instance, a damper stuck fault in an AHU can arise from multiple contributing causes, including a damper actuator power failure (e.g., power supply cause), disconnected communication wires of the damper controller (e.g., communication and network cause), and a mechanical failure of the linkage mechanism connecting the damper blades (e.g., hardware installation and configuration cause, or hardware degradation, age and fatigue cause). Without explicit representations of those contributing causes, it will be hard to enhance maintenance practice and improve equipment or system design.

In FDD-ON, we define this class as guiding FDD researchers and practitioners to carry out more investigations on understanding fault contributing causes. In the FDD-ON, the taxonomy of contributing causes are grouped into eight subclasses as listed in Table 7.

**Table** 7. List of contributing causes.

| Subclass | Annotation | Example of mapped faults |
|---|---|---|
| Manufacturing and design related cause (human) | The hardware or software are improperly designed and manufactured | Pipe is oversized |
| Hardware installation and configuration cause (human) | The component, equipment and system hardware are improperly installed and configured | Discharge air temperature location error fault |
| Software programming and configuration cause (human) | The software is improperly programmed and configured | PID setting fault |
| Hardware degradation, age and fatigue cause | The component, equipment and system hardware are degraded, aging and fatiguing during operations | Slippery fan belt fault |
| Unexpected external force on hardware cause | The component, equipment and system hardware suffers unexpected external force during operations | Damper leak fault |
| Environmental condition cause | The operational environment causes the hardware degradation | Coil fouling fault |
| Power supply cause | Improper or failure power supply | Damper obstruction fault |
| Communication and network cause | Improper and failure communication and network due to hardware or software configuration, interference, cyberattack | Thermostat IP setting fault |

**(4) Lifecycle occurrence stage**

The class of fault lifecycle occurrence stage attribute provides a structured temporal framework for identifying when the HVAC system' fault contributing causes originate and more importantly, when they can be systematically eliminated. By mapping fault types to these specific stages, an FDD system and AI-driven maintenance system can transition from merely detecting symptoms and diagnosing faults, to implementing targeted preventative strategies that eliminate faults at their point of origin. The sub class of attributes includes three individuals of lifecycle occurrence stage descriptions as listed in Table 8.

**Table** 8. List of Lifecycle occurrence stages.

| Subclass | Annotation | Example of mapped fault |
|---|---|---|
| Design stage | The fault originates from the system design | Pipe is oversized |
| Installation, commissioning, and service stage | The fault originates from when the system is installed, commissioned and served | Valve leak fault |
| Operation stage | The fault occurs when the system operates | Coil fouling fault |

### 3.2.5 Taxonomy of symptom attributes

The class of symptom attributes includes all the information that is required to characterize fault symptoms. The precise descriptions of symptom characteristics enable multiple applications. For example, many FDD approaches such as the rule-based approach and the Bayesian Networks approach rely on the fault symptom pattern identification to implement references of a fault. Currently, the class of symptom attributes includes two sub classes as: 1) symptom patterns, and 2) data types, as illustrated in Figure 3.

**(1) Symptom pattern**

Symptom patterns classify how symptom statuses are described when a fault occurs. In the development of FDD approaches, an identification and high-fidelity pattern matching fault symptom behaviors is essential to implement fault inference. Most existing FDD approaches rely on a simple steady state threshold-crossing approach to detect and diagnose faults. For example, if the supply air temperature of an AHU is constantly higher than the setpoint or the defined threshold, then a symptom (or abnormality) is detected and flagged by the FDD approach. Using this detected symptom and other symptoms, the FDD approach can infer that it is highly likely a cooling coil valve stuck fault. However, symptom behaviors can be more complex than a simple steady state threshold-crossing. Using the example above, the supply air temperature of an AHU can be oscillating, i.e., the supply air temperature may surpass the higher threshold and lower threshold frequently, indicating a fault in the controller (e.g., a PID setting fault). Very few research works have been conducted to uncover symptom patterns even though much research investigated using feature engineering approaches to extract or uncover symptom patterns [3]. Therefore, a classification of symptom patterns can enable the development of a more sophisticated, multi-dimensional pattern recognition of FDD approaches.

In FDD-ON, symptom patterns are classified into four subclasses i.e., level deviation pattern, temporal patterns, relation patterns, and statistical patterns as listed in Table 9. Under each subclass, a second level class to further describe the fault symptom patterns.

The subclass of the level deviation describes steady-state divergence, where a signal's value significantly surpasses from its expected baseline or reference values. For example, higher supply air temperature is classified into the 'High' second-level class under the 'Level deviation' subclass.

The subclass of temporal pattern characterizes the transient behavior of the symptom when fault occurs. This type of behavior captures anomalies in terms of how signal evolves over time. For instance, a slow response of a cooling coil valve position is classified into the 'Slow response' second-level class under the 'Temporal' subclass. This pattern can indicate the control parameter (i.e., the propagation gain) setting fault.

The subclass of relation pattern characterizes indicate that the symptom is described through multiple signals and manifested in terms of consistency and co-occurrence. For example, the outdoor air damper position and control signal mismatch symptom pattern uncovers an outdoor air damper stuck fault.

The subclass of statistical patterns uncovers the statistical characteristics of how symptoms manifest under specific fault types. For example, the symptom occurrence probability of discharge air temperature being too high can indicate the likelihood of the discharge air temperature abnormality under discharge air fan fault during a long-term operation [5].

**Table** 9. Classification of symptom patterns.

| Subclass | Annotation | Sub class (second level) |
| --- | --- | --- |
| Level deviation pattern | A symptom is quantified by the magnitude of deviation of measured signals from their expected values | High |
| | | Low |
| Temporal pattern | A symptom is defined by the temporal dynamics of system signals, including their transient and steady-state responses | Slow response |
| | | Unresponse |
| | | Oscillation |

| | | Cycling |
|---|---|---|
| | | Spike |
| | | Intermittent |
| Relation pattern | A symptom is described by relating to multiple signals and manifested in terms of consistency and co-occurrence. | Mismatch |
| | | Simultaneous Occurrence |
| Statistical pattern | A symptom is defined by the statistical properties of its status, such as occurrence probability and duration. | Occurrence probability |
| | | Continuous duration |

**(2) Data type**

Data types indicate what type of measurement(s) is (are) used to generate a symptom. Most FDD algorithms employ sensor data, control commands and settings, and operation state data, which are directly collected in the BAS to perform FDD. Some algorithms developed virtual sensors, which create derived data to enhance FDD performance [87]. In addition to BASs, some portable equipment (e.g., infrared camera [88]), which produces the multimedia data type, are used to monitor equipment operation. Lastly, an increasing number of FDD approaches and newly developed CMMS include metadata sources, such as historical maintenance service report data and fault alarming data, to enhance the capabilities of the fault diagnostics inference [89] and support the maintenance decision-making process [16].

Hence, to promote the interpretability and interoperability of symptom data across various systems, symptoms are categorized by their underlying data types in FDD-ON. In FDD-ON, the multidimensional taxonomy includes six data types as listed in Table 10.

**Table** 10. Classification of data types.

| Subclass | Annotation | Example of mapped symptoms |
|---|---|---|
| Sensor data | Data indicates sensor signals | Supply air temperature is high |
| Control command and setting | Data indicates from control commands and control settings in the BAS | Damper position control command is cycling |
| Operation state | Binary data indicates equipment operation state in the BAS | Chiller control is ON state |
| Text data | Text data indicates the occupant complaints; the operator notes, service logs, and system alarm descriptions | Occupant reports the zone temperature is high |
| Multimedia data | Data collected from multimedia devices (e.g., infrared camera) to indicate equipment operating conditions or abnormal behaviors | Discharge air fans have abnormal temperature distributions |
| Derived data | Data that is derived through analytics rather than directly observed | The relation of outdoor air damper position and control signal |

### 3.2.6 Taxonomy of impact types

The class of fault impact describes equipment-level or system-level impacts due to a specific fault. Chen et al. carried out a comprehensive review on HVAC and refrigeration (HVAC&R) system fault impacts from academic literature and industrial reports [3]. They reported that faults in HVAC&R systems cause impacts on equipment performance, energy, thermal comfort, indoor air quality, economic cost, maintenance activity and environment. FDD-ON adopts the classification of impacts and summarizes them into seven impact categories (i.e., subclasses) and 17 impact types (i.e., second level subclasses) listed in Table 11. It is noted that the CO2 rate was classified into the symptom class.

**Table** 11. Classification of fault impacts.

| Subclass (Impact category) | Second level subclass (Impact type) | Example of mapped faults |
|---|---|---|
| Equipment performance | COP | Refrigerant leakage |
| | EER | Refrigerant leakage |
| Energy | Electricity power | Cooling coil valve stuck |
| | Electricity power consumption | Cooling coil valve stuck |
| | Peak demand | Schedule error setting |
| | Gas consumption | Gas valve stuck |
| | Water consumption | Chilled water valve stuck |
| | Cooling | Chilled water valve stuck |
| | Heating | Heating valve stuck |
| Economic cost | Electricity energy consumption cost | Supply air temperature sensor bias |
| | Gas consumption cost | Gas valve stuck |
| | Water consumption cost | Chilled water valve stuck |
| | Maintenance money | Heating valve stuck |
| Indoor air quality | Outdoor air ratio | Discharge air damper stuck |
| Thermal comfort | Predicted percentage of dissatisfied | Discharge air damper stuck |
| Maintenance activity | Maintenance time | Supply air fan motor failure |
| Environment | Emission | Refrigerant leakage |

### 3.2.7 Implementation of semantic modeling and ontology visualization

In this study, we translate the semantic model of FDD-ON into OWL and RDFS using Protégé software [90]. Protégé provides a GUI to allow the identification and creation of classes, class hierarchies, variables, and the relationships between classes and the properties of these relationships. Figures 4 and 5 show the developed hierarchy in the Protégé tool.

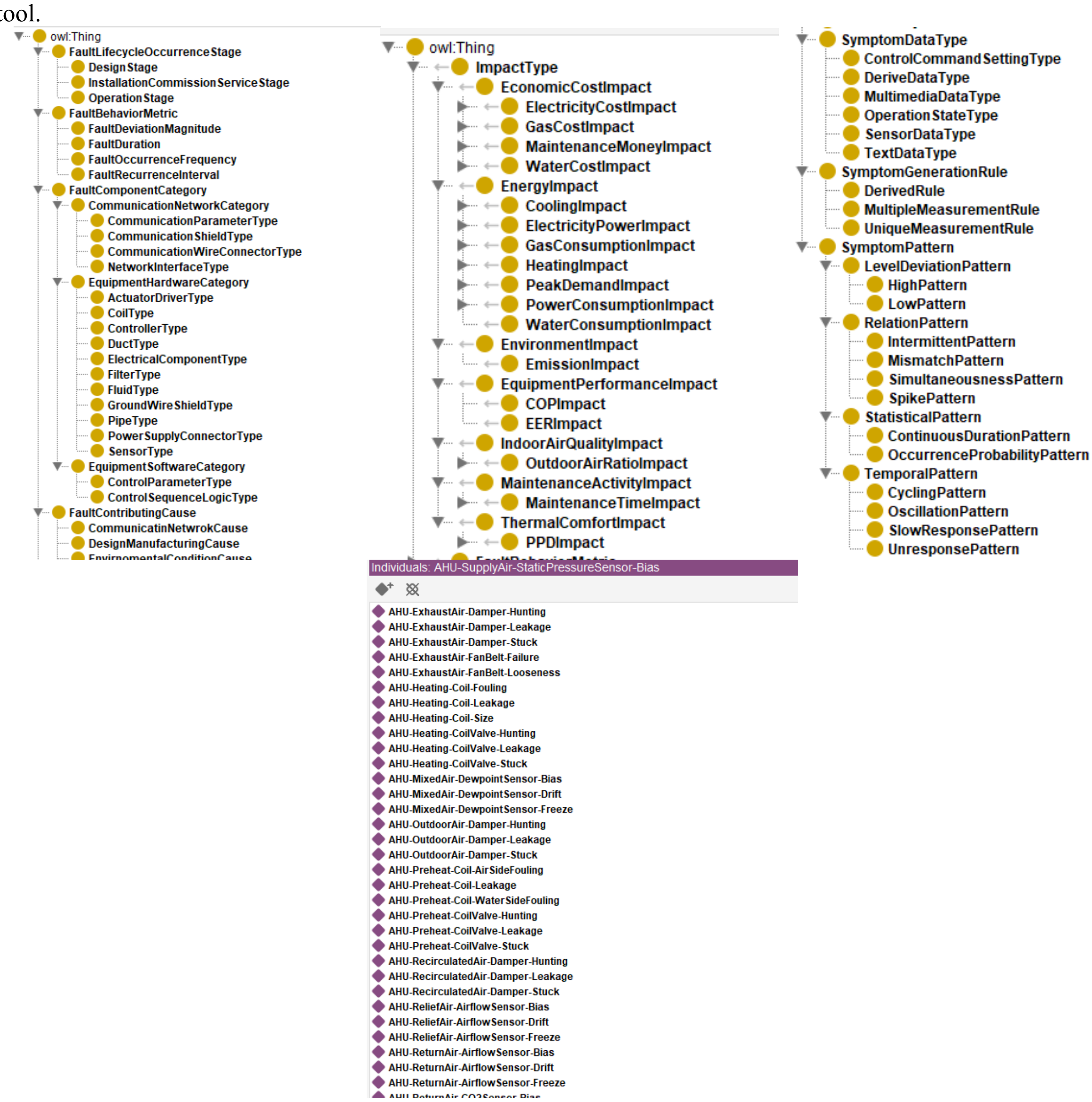

**Figure** 4. Example hierarchy of the classes for fault attributes, impact types, symptom attributes and individuals in the Protégé tool.

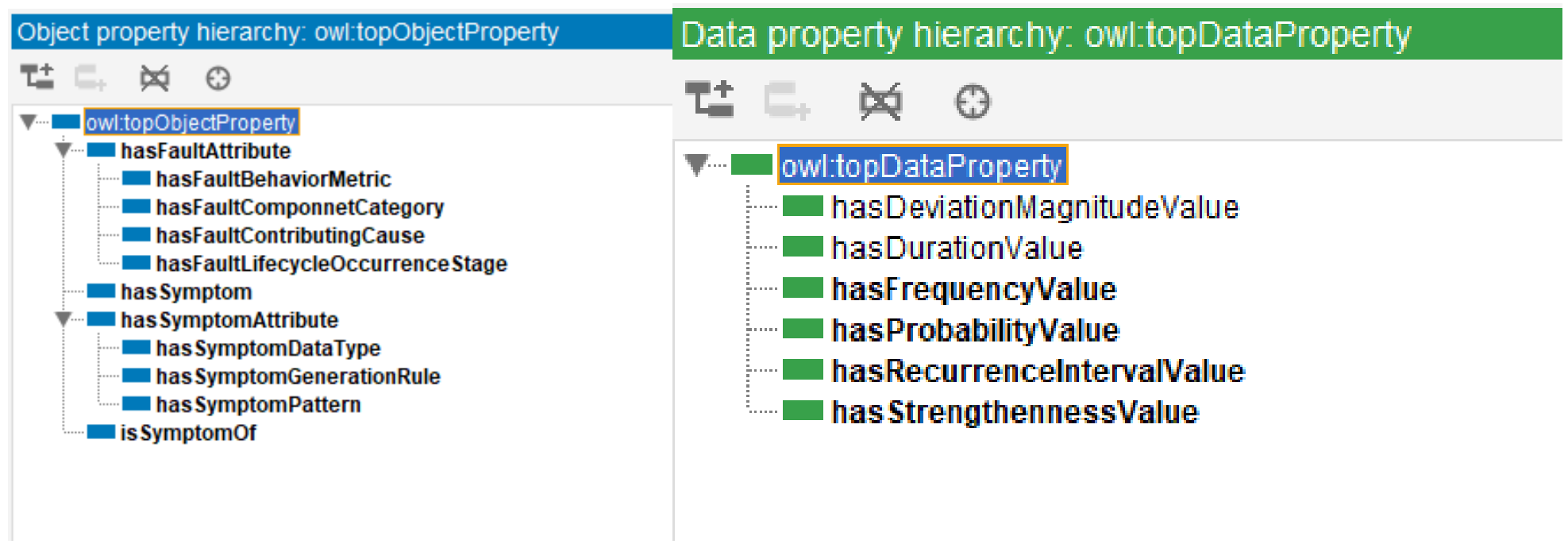

**Figure** 5. Example hierarchy of Properties in the Protégé tool.

Figures 6 and 7 show examples of the graphic models of the defined fault types (i.e., '*VAVATU-DischargeAir-AirflowSensor*' related faults) and symptoms (i.e., the abnormalities manifesting on the measurement of '*VAVATU-DischargeAir-Airflow*') in the FDD-ON in the fault library and symptom library. The graphic representations explicitly demonstrate fault types and associated symptoms, as well as the aligned attributes.

Under the '*VAVATU-DischargeAir-AirflowSensor*' fault subclass, it includes four specific fault types indicated by (the purple heart symbols in Figure 6: bias, drift, freeze, and error location. The fault subclass has all four types of fault behavior metric as defined in Table 5. The component where the fault occurs is the airflow sensor. So, it is classified into the '*SensorType*' under the '*EquipmentHardewareCategory*' subclass. The fault has five contributing causes: '*PowerCause*', '*DesignManfacturingCause*', '*HardwareInstallationConfigurationCause*', '*HardwareDegradationAageFatigueCause*', and '*CommunicationNetworkCause*'. The faults often occur at the installation and commission service stage and operational stage, so it is classified in the '*InstallationCommissionServiceStage*' and '*OperationStage*' in terms of fault lifecycle occurrence stage. Additionally, the faults may cause symptoms on ten measurements. They are '*VAVATU-DischargeAir-Temperature*', '*VAVATU-DischargeAir-Airflow*', '*VAVATU-DischargeAir-DamperControl*', '*VAVATU-DischargeAir-DamperPosition*', '*VAVATU-DischargeAir-Dewpoint*', '*VAVATU-Reheat-CoilValve-Control*, '*VAVATU-Reheat-CoilValve-Position*', '*VAVATU-SupplyDischargeAir-Temperture*', '*VAVATU-Thermostat-Temperature*', '*VAVATU-DischargeAir-FanSpeed*', and '*VAVATU-DischargeAir-FanSpeedControl*'. Additionally, the faults have impacts in terms of 'Cooling', 'Heating', 'ElectricityPower', 'PowerConsumptojn''ElectricityCost'

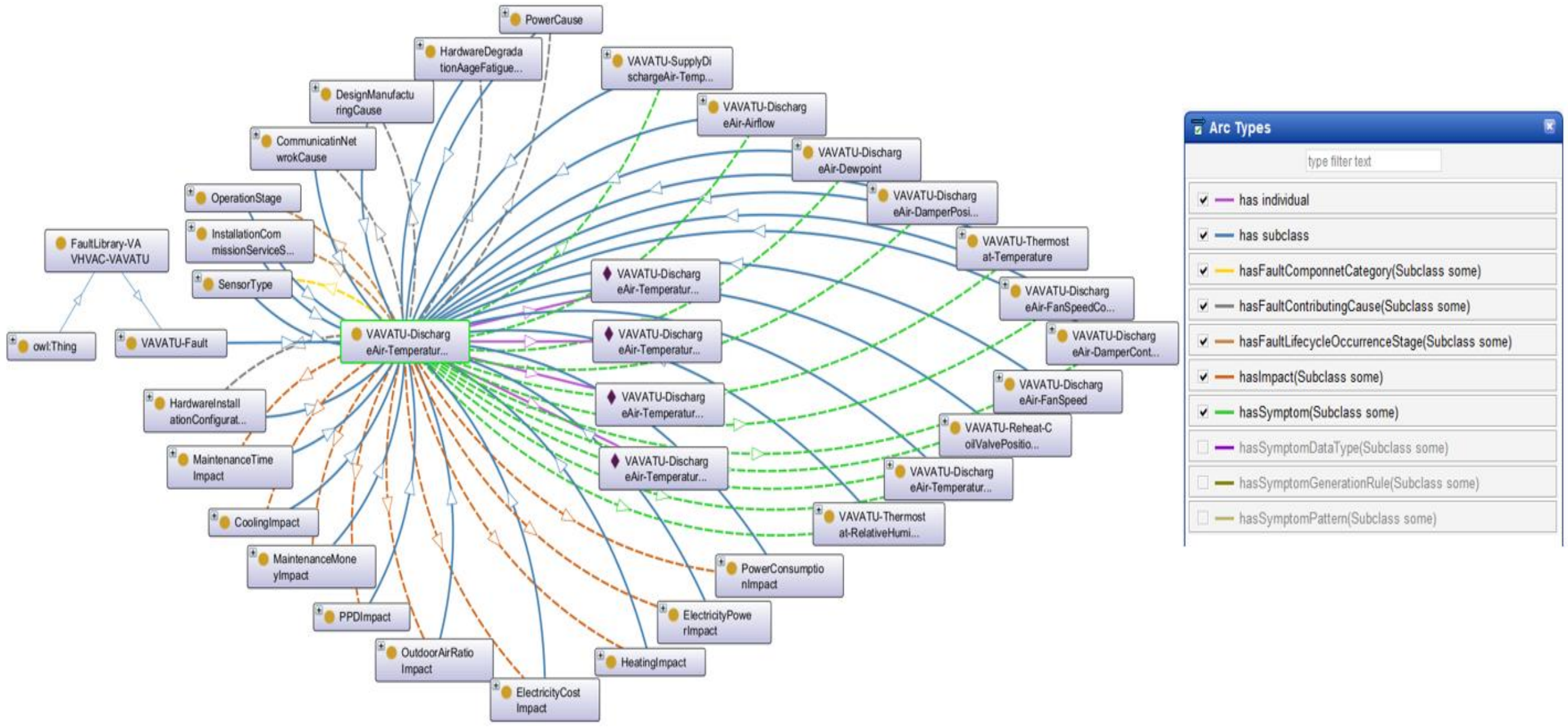

**Figure** 6. Example of an FDD-ON graphic model of VAV-ATU fault types (VAVATU-DischargeAir-AirflowSensor faults).

Under the VAVATU discharge air airflow abnormal symptom, it includes five specific fault statuses (i.e., high, low, un-response, oscillation and slow response). This measurement is classified into the 'SensorDataType' because the symptom is manifested by the air airflow sensor. The five statuses are classified into three patterns as 'LevelDeviationPattern', 'Temporal pattern' and 'StatisticalPatten'. The symptom is generated by a unique measurement (i.e., one sensor). The symptoms can be caused by 26 classes of faults in the VAVATUs.

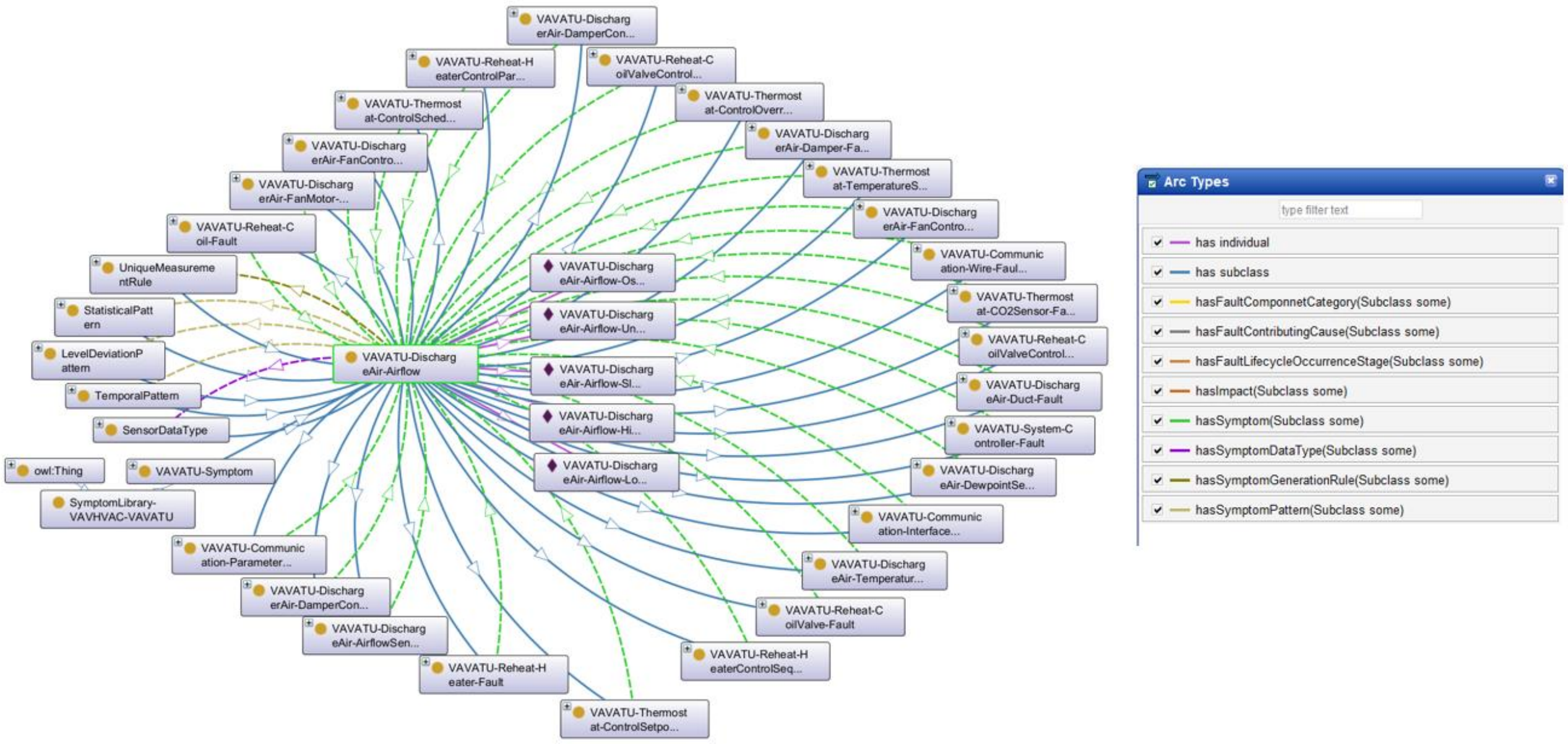

**Figure** 7. Example of an FDD-ON graphic model of VAV-ATU symptom statuses (VAVATU-DischargeAir-Airflow abnormal).

### 3.3 Development of libraries

#### 3.3.1 Naming convention

The first step to develop the fault library and symptom library is to define naming conventions in ontology. In this study, we follow the rules to represent both a specific fault and a specific type of symptom as developed [37]. However, a major modification compared to [37] is that most behavior-based faults are defined fault symptoms to ensure a clear causal-effect relation and to guarantee robust semantic-based inference.

There are two primary considerations when developing the naming convention of fault types and symptoms.

First, the naming convention should be concise and accurate to capture the natures of fault types and symptom statuses, e.g., the equipment and component locations where a fault occurs, and the measurement types that manifest symptoms. This enhances the interpretability of the data; and hence, reduces ambiguities when developing FDD related solutions, as well as facilitating interpreting and translating FDD results into actionable maintenance activities. Therefore, In the fault names and symptom names, we try to avoid using the abbreviations of component types, locations and descriptions of fault types and symptom statuses. Instead, we only use abbreviations for equipment types, e.g., we use AHU to in the name to represent that a fault type or a symptom status is associated with an air handling unit.

Secondly, a robust naming structure should be adopted to reduce the heterogeneity of the data, and consequently increase the data interoperability, as well as efficient embedding and processing across different system environments. All fault names, symptom names and impact names are constructed into a four-section structure, namely a **quadrinomial nomenclature**. The four-section structure encodes fundamental information to illustrate a fault type, symptom status and impact type. In each section, Pascal Case notations [91] are adopted to represent specific pieces of information that are associated with characteristics of faults, symptoms and impact types. It is noted that we adopted hyphens to connect each section in the quadrinomial nomenclature because using hyphens as separators in Uniform Resource Locator (URLs) can be easily read. However, underscores can also be used to connect sections when programming the names in computer languages.

For a specific fault name, the four-section structure is constructed by: i) the equipment where the fault occurs; ii) the location of the equipment where the faulty component is located; iii) the physical component where the fault occurs; and iv) the controlled vocabulary that specifies a fault mode, respectively. Table 12 exemplifies a fault name using the defined naming convention. The full name of a damper leakage fault occurring at the outdoor air section of an AHU is hereby named as '*AHU-OutdoorAir-Damper-Leakage*'.

**Table** 12. Illustration of a standardized fault naming convention

| **Section description** | **Equipment** | **Information specifying the location** | **Information specifying the component type** | **Controlled vocabulary specifying the fault mode** |
|---|---|---|---|---|
| Example | AHU | OutdoorAir | Damper | Leakage |

Similarly, for a specific symptom name, the four-section structure encodes information, which represents i) the equipment where the fault occurs; ii) the location of the measurement where the fault symptom is measured; iii) the measure that manifests the symptom; and iv) the controlled vocabulary, specifies a symptom status. Table 13 exemplifies a symptom name using the defined naming convention. The full name of a symptom describing an AHU supply air temperature is high is hereby given as '*AHU-SupplyAir-Temperature-High*'.

**Table** 13. Illustration of a standardized symptom naming convention

| **Section description** | **Equipment** | **Information specifying the location of equipment** | **Measurement type** | **Controlled vocabulary specifying the symptom status** |
|---|---|---|---|---|
| Example | AHU | SupplyAir | Temperature | High |

The naming convention for  impacts has the same three sections defined in the naming convention for the fault type, i.e., equipment type, information specifying a location of equipment, and information specifying a component type. The fourth section defines the impact type. Table 14 exemplifies a symptom name using the defined naming convention. The full name of '*AHU-SupplyAir-FanMotor-PowerConsumption*' indicates the impact of an AHU supply air fan motor in terms of power consumption.

**Table** 14. Illustration of a standardized symptom naming convention

| **Section description** | **Equipment** | **Information specifying the location of equipment** | **Information specifying the component type** | **Impact type** |
|---|---|---|---|---|
| Example | AHU | SupplyAir | FanMotor | PowerConsumption |

### 3.3.2 Description of libraries

FDD-ON includes three major libraries, i.e., fault library, symptom library and impact library for each type of subsystem and equipment. Currently, there are 469 fault types, 468 symptom statuses, and 447 impact types across five equipment types, which have been included in the fault libraries, symptom libraries and impact libraries, respectively. These fault types, symptom statuses and impact types have been reported in 241 specific components and 137 measures, respectively. Table 15 lists the distributions of fault types, symptom statuses, and impact types (i.e., entities) in each type of subsystem and equipment.

**Table** 15. Entity distributions

| Equipment type | Number of component types | Number of fault types | Number of measures | Number of symptom statuses | Number of impact types |
|---|---|---|---|---|---|
| Chiller plant | 42 | 85 | 25 | 91 | 82 |
| Boiler plant | 19 | 37 | 10 | 37 | 38 |
| AHU | 68 | 130 | 55 | 176 | 118 |
| RTU | 83 | 164 | 33 | 114 | 160 |
| VAV ATU | 29 | 537 | 14 | 50 | 49 |
| Total | 241 | 469 | 137 | 468 | 447 |

## 4. Evaluation and discussion

The ontology evaluation is a critical step in ontology engineering to ensure the developed model is logically sound, accurately represents the target domain, and effectively fulfills its intended application requirements [92,93]. We evaluated FDD-ON from three angles. First, we illustrate the quantitative evaluation to demonstrate the structural quality and completeness in Section 4.1. Secondly, we explain the assessment of logical consistency in Section 4.2. Thirdly, we showcase the application of FDD-ON to demonstrate its application in FDD development in Section 4.3. At the same time, we discuss the limitations in the current version of FDD-ON in Section 4.4 and we discuss the evolution directions of FDD-ON in Section 4.5.

### 4.1 Quantitative evaluation

#### 4.1.1 Ontology metrics

We evaluate the structural quality by analyzing the entities used in the current version of FDD-ON, as listed in Table 16.

In the table, the axiom counts of 8,756 represents the formal statement of truth within FDD-ON by including class declarations, subclass relationships, property domains, and disjointness rules. The 6,537 logical axiom counts exclude purely structural or decorative statements (like metadata annotations, labels, and comments) and counts only the assertions that the DL reasoner evaluates to infer new knowledge. The 2,188 declaration axioms count illustrates the number of times that a brand-new entity (a new class, property, or individual) is introduced to the FDD-ON. The entity inventory (including class count, object property count and named individual count) indicates that FDD-ON consists of 775 distinct subclasses organized to represent VAV HVAC system fault types, symptom statues, impact types and associated attributes. Semantic richness is introduced via 15 object properties, 7 data properties and 1,393 individuals.

**Table** 16. Ontology metrics.

| Name | Number of entities |
|---|---|
| Axiom | 8,756 |
| Logical axiom count | 6,537 |
| Declaration of axioms count | 2,188 |

| Class count | 775 |
|---|---|
| Object property count | 15 |
| Data property count | 7 |
| Individual count | 1,393 |

### 4.1.2 Completeness evaluation

We evaluate the information completeness of FDD-ON by analyzing existing publicly available FDD data sets for VAV HVAC systems. The data sets were generated using real systems, lab tests and software simulations for the purpose of developing and benchmarking FDD approaches and other applications. Hence, the data sets include common fault types and measurements in VAV HVAC systems. The data sets cover the chiller plant subsystem, boiler plant subsystem, AHUs, RTUs and ATUs (including both hydronic reheat ATUs and fan power units). Table 17 lists the data mapping results.

**Table** 17. Data mapping analysis.

| Dataset Name | Equipment type | Number of component types | Number of fault types | Pct of component type mapped | Pct of fault types mapped | Number of measurement types | Number of measurement types mapped | Pct of measurements mapped |
|---|---|---|---|---|---|---|---|---|
| Drexel-Nesbitt [69] | AHU | 4 | 4 | 100% | 100% | 21 | 17 | 81% |
| Drexel-Nesbitt [69] | Chiller | 4 | 4 | 100% | 100% | 25 | 25 | 100% |
| Drexel-Nesbitt [69] | VAV ATU | NA | NA | NA | NA | 5 | 5 | 100% |
| ASHRAE1312 [94] | AHU | 8 | 10 | 100% | 100% | 43 | 35 | 81% |
| LBNL dataset [95] | Dual duct AHU | 11 | 16 | 100% | 100% | 54 | 51 | 94% |
| LBNL dataset [95] | FPU | 8 | 10 | 100% | 100% | 16 | 16 | 100% |
| LBNL dataset [95,96] | RTU | 11 | 16 | 100% | 100% | 22 | 21 | 95% |
| LBNL dataset [95] | Chiller plant | 6 | 7 | 100% | 100% | 38 | 37 | 97% |
| LBNL dataset [95] | Boiler plant | 5 | 5 | 100% | 100% | 13 | 13 | 100% |
| LBNL dataset [96] | AHU (multizone)-1 | 4 | 4 | 100% | 100% | 17 | 16 | 94% |
| LBNL dataset [96] | AHU (singlezone) | 3 | 5 | 100% | 100% | 14 | 13 | 93% |
| LBNL dataset [95] | AHU (multizone)-2 | 4 | 5 | 100% | 100% | 25 | 24 | 96% |
| ORNL dataset [97] | VAV ATU | 2 | 2 | 100% | 100% | 7 | 6 | 86% |
| ORNL dataset [97] | RTU | NA | NA | NA | NA | 15 | 15 | 100% |

It can be seen that 100% fault types in each FDD data set were mapped in the fault library of the FDD-ON. For the measurement types, the unmapped measurement types primarily consist of two types of data: 1) meter data and 2) equipment status data. Meter data were not considered in the symptom library because those data were categorized into impact types. Some equipment status data and BAS data points (e.g., the data points 'System occupancy control mode' and 'Chiller selection' in ASHRAE RP1312 dataset) were not included in the symptom library, and hence the corresponding measurements were considered as unmapped measurement types.

### 4.2 Qualitative evaluation

We carried out a consistency examination to assess logical expressivity. During the development of FDD-ON, DL examinations were frequently run using the Pellet plug-in in the Protégé software to ensure the DL consistency of the FDD-ON. The reasoner engine uses rules and axioms to check unsatisfiable classes (e.g., the intersection errors), property domain and range checking, and other DL errors. FDD-ON successfully passes the consistency check.

### 4.3 Application evaluation

#### 4.3.1 Feature comparison

We compare the semantic modelling approach of FDD-ON with other ontologies to demonstrate the enhancement of FDD-ON in representing FDD domain knowledge and application areas.

Appendix I lists the feature comparison between FDD-ON and other ontologies in buildings to highlight the characteristics of FDD-ON. The comparison demonstrates that FDD-ON provides a more comprehensive knowledge representation for faults, symptoms and impacts in VAV HVAC systems than existing ontologies or semantic models. Therefore, the features in FDD-ON sufficiently address the CQs proposed in Section 3.1.2 and can be employed in various applications.

#### 4.3.2 Example of an application

(1) Description of the FDD dataset

We use an FDD dataset, which was constructed by real operational data in an office building to demonstrate the application of FDD-ON. The dataset was published in [69]. In the dataset, fault inclusive data and fault free data were collected via the BAS in the Nesbitt Hall, in Philadelphia, PA, U.S. The building is a seven-story, 7,260 $m^2$ mixed-use commercial building. The HVAC system includes a water-cooled chiller plant subsystem and one boiler plant subsystem, which provides primary cooling, heating and domestic hot water for the building. The air distribution subsystem includes three single-duct AHUs and eighty-eight VAV ATUs to condition zones in the building's floors. The HVAC system is monitored and controlled by a BAS.

(2) Fault mapping result

The FDD dataset includes eight types of faults manually implemented in the chiller plant and AHU as listed in Table 18. The table provides the original fault descriptions, fault severity levels, fault durations, and mapped fault names in the FDD-ON fault library. It can be seen that fault descriptions are 100% mapped to fault names in FDD-ON. The two pieces of information (i.e., fault severity levels and fault durations) can be mapped into '*FaultDeviationMagnitude*' and '*FaultDuration*' in the '*FaultBehaviorMetric*' subclass in FDD-ON.

**Table** 18. Mapped fault types in [69].

| Subsystem/ Equipment | Fault description | Fault severity | Fault duration (minutes) | Mapped fault name in FDD-ON |
|---|---|---|---|---|
| Chiller plant | Occupied during the unoccupied schedule;<br>Chiller stopped earlier than scheduled;<br>HVAC system occupied earlier than scheduled;<br>Change weekend occupied schedule to end | NA | 420; 690; 240; 80 | CHILLERPLANT-Control-Schedule-SettingError |
| | System stopped working (Chiller off) | NA | 120 | CHILLERPLANT-System-Controller-Offline |
| | Chiller supply chilled water temperature sensor bias | −2.3°C;<br>−1.7°C | 687; 600 | CHILLERPLANT-PrimaryLoopChilledWaterSupply-TemperatureSensor-Bias |
| | Chiller chilled water differential pressure sensor positive bias | 0.69 kPa | 291 | CHILLERPLANT-SecondaryLoopChilledWater-DifferentialPressureSensor-Bias |
| AHU | AHU outdoor air damper stuck | 30%; 60%; 80%, 90%; 100% | 480; 600; 630; 645, 601; 541 | AHU-OutdoorAir-Damper-Stuck |
| | AHU cooling coil valve position software override | 100% | 601 | AHU-OutdoorAir-DamperControlSequence-SettingError |
| | AHU cooling coil valve stuck | 80% | 600 | AHU-Cooling-CoilValve-Stuck |
| | AHU supply air temperature sensor bias | −2°C | 480 | AHU-SupplyAir-TemperatureSensor-Bias |

(3) Symptom and impact mapping result

The FDD dataset includes more than 500 data points (measurements) collected from the BAS. The original data points cover three data types, i.e., sensor data type, control command and control setting type, and discrete operation state type. However, the measurement can also produce the derived data type after analyzing the data. For example, a symptom can be produced by comparing the outdoor air temperature and mixed air temperature when diagnosing the outdoor damper related faults [54]. Due to the length of the paper, we listed part of the mapped symptoms, as well as the corresponding data types in Table 19.

**Table** 19. Example of mapped symptom statuses in [69].

| Data point name | Description | Mapped symptom name in FDD-ON | Data type |
|---|---|---|---|
| #drx_nsbt_chws_control/chw_flow | Sensor: Chilled water flowrate | CHILLERPLANT-PrimaryLoopChilledWaterSupply-FlowRate-High | LevelDeviation |
| | | CHILLERPLANT-PrimaryLoopChilledWaterSupply-FlowRate-Low | LevelDeviation |
| | | CHILLERPLANT-PrimaryLoopChilledWaterSupply-FlowRate-UnResponse | TemporalPattern |
| | | CHILLERPLANT-PrimaryLoopChilledWaterSupply-FlowRate-Oscillation | TemporalPattern |
| #drx_nsbt_chws_control/chiller_status | Control chiller status | CHILLERPLANT-Chiller-ControlStatus-Mismatch | RelationPattern |
| #drx_nesbitt_ahu-2/sf_vfd_output/trend_log | Sensor: Supply fan speed | AHU-SupplyAir-FanSpeed-High | LevelDeviation |
| | | AHU-SupplyAir-FanSpeed-Low | LevelDeviation |
| | | AHU-SupplyAir-FanSpeed-Unresponse | TemporalPattern |
| | | AHU-SupplyAir-FanSpeed-Oscillation | TemporalPattern |
| #drx_nesbitt_vv-1-b-2_amd33/damper_pos | Control: Damper position | VAVATU-DischargeAir-DamperControl-Mismatch | RelationPattern |
| | | VAVATU-DischargeAir-DamperControl-Cycling | TemporalPattern |

(4) Knowledge querying for FDD construction

Knowledge querying is one of the important activities when embedding ontology technologies into various applications. When developing FDD approaches, knowledge querying provides a mechanism that an AI or inference engine uses to extract context, construct cause-effect structures, and perform automated reasoning. Since ontologies have interpretable data, structured data formats, and complete relations, knowledge querying becomes efficient to leverage both explicitly stored knowledge and inferred knowledge generated by reasoners.

There are several approaches such as SPARQL (Semantic Protocol and RDF Query Language), DL Queries and Semantic Web Rule Language (SWRL) that can be used to implement semantic search and data query. The definition of query protocol of SPARQL can be found in the [98].

Figure 8 illustrates the SPARQL clause and the result after executing SPARQL search in Protégé. This shows the measurements which may manifest symptoms associated with '*AHU-SupplyAir-TemperatureSensor*' faults (including three types as bias, drift and freeze). The faults can produce symptoms on 11 measurements (i.e., symptom subclass in Figure 8(b)). Figure 9 illustrates a graphic visualization of the attributes and symptom subclasses associated with this fault subclass. It can be seen that the knowledge querying result can be used to construct certain diagnostics structures, such as rule-based methods and Bayesian Network (BN) methods, where causal-effect relations should be embedded. Additionally, DT technology can be efficiently developed with the unified and interoperable data structure provided by FDD-ON.

```
PREFIX rdf: <http://www.w3.org/1999/02/22-rdf-syntax-ns#>
PREFIX owl: <http://www.w3.org/2002/07/owl#>
PREFIX rdfs: <http://www.w3.org/2000/01/rdf-schema#>
PREFIX xsd: <http://www.w3.org/2001/XMLSchema#>

PREFIX fdd: <http://www.semanticweb.org/y6g/ontologies/2025/FDD_ON_VAV_HVAC#>
SELECT ?symptomClass
WHERE {
 BIND(fdd:AHU-SupplyAir-TemperatureSensor-Fault AS ?faultClass)

 ?faultClass rdfs:subClassOf ?restriction .
 ?restriction rdf:type owl:Restriction .
 ?restriction owl:onProperty fdd:hasSymptom .

  { ?restriction owl:someValuesFrom ?symptomClass . }
 UNION
 { ?restriction owl:allValuesFrom ?symptomClass . }
}
```

| symptomClass |
| --- |
| AHU-SupplyAir-StaticPressureSetpoint |
| AHU-SupplyMixAir-Temperature |
| AHU-SupplyAir-StaticPressure |
| AHU-Cooling-CoilValvePosition |
| AHU-Heating-CoilValvePosition |
| AHU-SupplyAir-Temperature |
| AHU-Heating-CoilFlowRate |
| AHU-Cooling-CoilFlowRate |
| AHU-Cooling-CoilLeavingTemperature |
| AHU-Heating-CoilLeavingTemperature |
| AHU-SupplyAir-TemperatureSetpoint |

(a) SPARQL clause (b) Queried symptom subclass

**Figure** 8. Example of SPARQL clause and the knowledge querying result

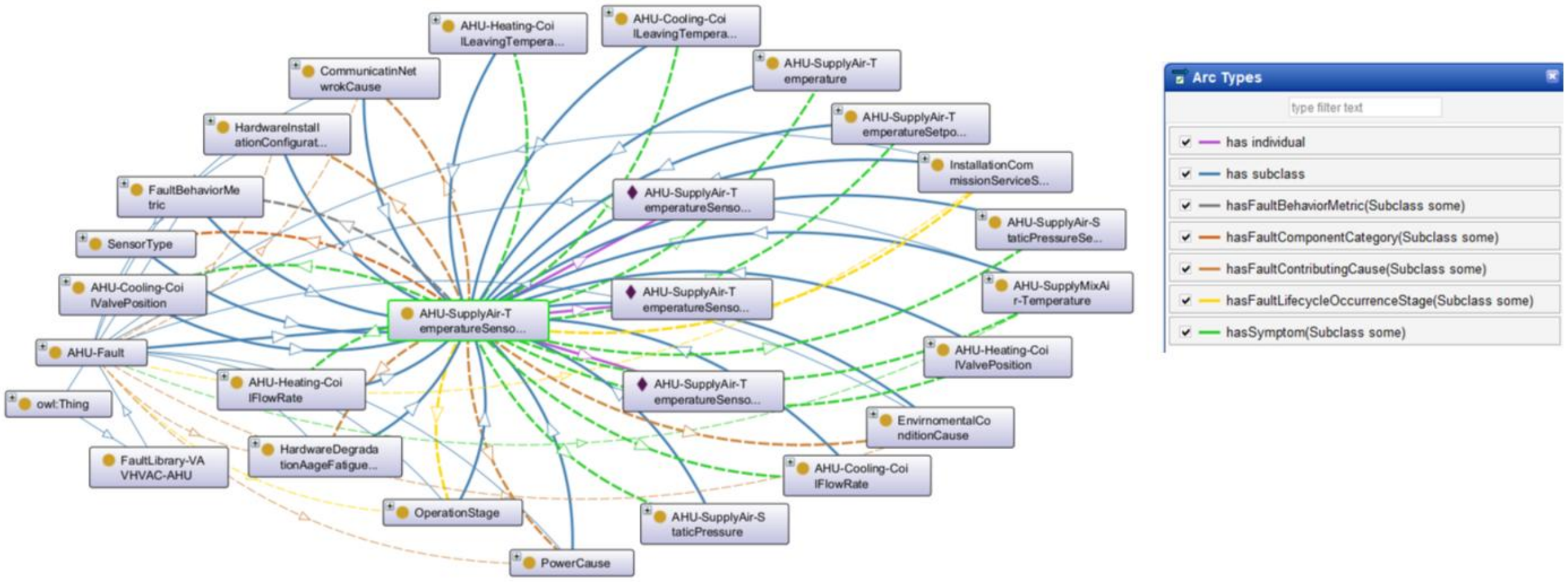

**Figure** 9. Graphic visualization of the attributes and symptom subclasses associated with ''*AHU-SupplyAir-TemperatureSensor*' fault subclass.

### 4.4 Limitations and ontology evolution

The development of ontologies is an evolving and iterative process because it requires new incoming knowledge and practical experience during practices. The current FDD-ON has two major limitations in terms of the knowledge representations and data structure as:

> (1) We only developed one-level taxonomy to categorize fault contributing causes, which may be insufficient to identify contributing causes at a more granularity level. Additionally, we did not consider the fault propagations, i.e., a fault can be a contributing cause for other types of faults. Although this simplifies the current relations between fault contributing causes and faults, it may limit more complicated analytics.
>
> (2) In the current FDD-ON, the relations between faults and symptoms, and the relations between faults and impacts are restricted to the equipment level, that's said, no cross-level casual effect relations among faults, symptoms and impacts were developed. This may limit the development of the FDD approach and perform fault impact analyses.

In addition to addressing the limitations, the research team plans to update and evolve FDD-ON from three angles.

(1) We will evolve FDD-ON by adding more equipment types, fault types, as well as associated symptoms and impacts in the HVAC systems. In addition, we will expand the representation of faults and symptoms in HVAC systems with the addition of new terms as needed for data curation and collaborative development.

(2) We will investigate more fault mechanisms and symptom statuses and then expand the controlled vocabulary list.

(3) We will study and include the knowledge related to maintenance activities into FDD-ON.

## 5. Conclusion

Efficient O&M for VAV HVAC systems are essential for ensuring energy efficiency, improving indoor comfort, reducing operational maintenance costs and extending equipment service lifespan. Achieving these objectives relies on the integration of multiple advanced technologies, among which, FDD plays a critical role. Successful deployment and implementation of FDD requires not only robust algorithms capable of accurately identifying abnormalities and diagnosing faults but also needs structured data representations and semantic models that support DT-based FDD applications, AI-driven maintenance decision-support systems, and fault tolerant control strategies. The integration of these technologies enables a closed-loop process that connects fault detection, diagnosis, corrective actions, and impact mitigation, thereby enhancing the overall reliability, efficiency, and sustainability of HVAC system operations. However, there is a lack of ontology targeting to cover the knowledge richness and promote FDD knowledge sharing for VAV HVAC systems in existing practice.

In this study, we develop an ontology – FDD-ON, which employs knowledge representation and semantic modeling, to enhance data interpretability and semantic interoperability, as well facilitate knowledge sharing across various applications for VAV HVAC systems. We empower FDD-ON with several novel features, aiming at providing an open source and formal ontological representation of faults and symptoms in VAV HVAC systems:

(1) We consider the requirements of multidisciplinary environments and applications, which FDD can be integrated into, when developing the four-level semantic model.

(2) We develop various taxonomies to classify various attributes associated with faults and symptoms to completely represent information at different levels related to FDD and hence enhance the semantic reasoning.

(3) We constructed an open-source fault library and a symptom library, which provide the unified definitions of fault types and associated symptom statuses in HVAC systems. Controlled vocabulary items are developed to unify the semantic representations of fault types and symptom statuses so that data interpretability can be enhanced. The standardized and structured naming conventions for fault types and symptom statuses enhance the data interoperability, making it suitable to various applications. To date, the FDD-ON includes 469 fault types and 468 symptom statuses across 241 types of components and 137 measures, as well as 447 impact types in VAV HVAC systems.

(4) We evaluate FDD-ON from multiple angles to show that the existing version of FDD-ON well captures the information in the FDD area. In addition, the nature of FDD-ON enables its evolution to address new gaps in different applications.

In the future, in addition to addressing the limitations and evolving FDD-ON as discussed, we will showcase more applications of FDD-ON in terms of DT deployment, AI-driven maintenance-supporting system development and fault tolerant control strategies.

## References

[1] M. González-Torres, L. Pérez-Lombard, J.F. Coronel, I.R. Maestre, D. Yan, A review on buildings energy information: Trends, end-uses, fuels and drivers, Energy Rep. 8 (2022) 626–637. https://doi.org/10.1016/j.egyr.2021.11.280.

[2] Grand View Research, HVAC Maintenance Services Market Size, Share & Trends Analysis Report (2024–2030), 2025. https://www.grandviewresearch.com/industry-analysis/hvac-maintenance-services-market-report (accessed January 1, 2026).

[3] Y. Chen, Y. Hu, G. Lin, Y. Zhang, S. Ye, B. Shen, Evaluation of HVAC & refrigeration system fault behaviors and impacts: A systematic review, J. Build. Eng. (2025) 113609. https://doi.org/10.1016/j.jobe.2025.113609.

[4] S. Katipamula, M.R. Brambley, Methods for Fault Detection, Diagnostics, and Prognostics for Building Systems—A Review, Part I, HVACR Res. 11 (2005) 3–25. https://doi.org/10.1080/10789669.2005.10391123.

[5] Y. Chen, Z. Chen, G. Lin, J. Wen, J. Granderson, A Simulation-Based Method to Analyze Fan Coil Unit Fault Impacts, in: ASHRAE Trans., ASHRAE, Toronto, Canada, 2022: p. 210. https://www.ashrae.org/technical-resources/ashrae-transactions.

[6] S. Katipamula, R.M. Underhill, N. Fernandez, W. Kim, R.G. Lutes, D. Taasevigen, Prevalence of typical operational problems and energy savings opportunities in U.S. commercial buildings, Energy Build. 253 (2021) 111544. https://doi.org/10.1016/j.enbuild.2021.111544.

[7] E. Crowe, Y. Chen, H. Reeve, D. Yuill, A. Ebrahimifakhar, Y. Chen, L. Troup, A. Smith, J. Granderson, Empirical analysis of the prevalence of HVAC faults in commercial buildings, Sci. Technol. Built Environ. 29 (2023) 1027–1038. https://doi.org/10.1080/23744731.2023.2263324.

[8] Z. Chen, Z. O'Neill, J. Wen, O. Pradhan, T. Yang, X. Lu, G. Lin, S. Miyata, S. Lee, C. Shen, R. Chiosa, M.S. Piscitelli, A. Capozzoli, F. Hengel, A. Kührer, M. Pritoni, W. Liu, J. Clauß, Y. Chen, T. Herr, A review of data-driven fault detection and diagnostics for building HVAC systems, Appl. Energy 339 (2023) 121030. https://doi.org/10.1016/j.apenergy.2023.121030.

[9] Y. Yu, D. Woradechjumroen, D. Yu, A review of fault detection and diagnosis methodologies on air-handling units, Energy Build. 82 (2014) 550–562. https://doi.org/10.1016/j.enbuild.2014.06.042.

[10] W. Kim, S. Katipamula, A review of fault detection and diagnostics methods for building systems, Sci. Technol. Built Environ. 24 (2018) 3–21. https://doi.org/10.1080/23744731.2017.1318008.

[11] Z. Shi, W. O'Brien, Development and implementation of automated fault detection and diagnostics for building systems: A review, Autom. Constr. 104 (2019) 215–229. https://doi.org/10.1016/j.autcon.2019.04.002.

[12] J. Bi, K. Yan, Y. Du, End-to-end residual learning embedded ACWGAN for AHU FDD with limited fault data, Build. Environ. 270 (2025) 112529. https://doi.org/10.1016/j.buildenv.2025.112529.

[13] J. Schein, S.T. Bushby, N.S. Castro, J.M. House, A rule-based fault detection method for air handling units, Energy Build. 38 (2006) 1485–1492. https://doi.org/10.1016/j.enbuild.2006.04.014.

[14] T. Li, Y. Zhao, C. Zhang, J. Luo, X. Zhang, A knowledge-guided and data-driven method for building HVAC systems fault diagnosis, Build. Environ. 198 (2021) 107850. https://doi.org/10.1016/j.buildenv.2021.107850.

[15] S. Frank, G. Lin, X. Jin, R. Singla, A. Farthing, J. Granderson, A performance evaluation framework for building fault detection and diagnosis algorithms, Energy Build. 192 (2019) 84–92. https://doi.org/10.1016/j.enbuild.2019.03.024.

[16] Y. Chen, E. Crowe, G. Lin, J. Granderson, Integration of FDD data to aid HVAC system maintenance, in: Proc. 9th ACM Int. Conf. Syst. Energy-Effic. Build. Cities Transp., Association for Computing Machinery, New York, NY, USA, 2022: pp. 492–495. https://doi.org/10.1145/3563357.3567405.

[17] H.M. Newman, BACnet: The Global Standard for Building Automation and Control Networks, Momentum Press, 2013.

[18] N. Kruis, T. McDowell, C. S. Barnaby, T. Mankad, ASHRAE Standard 205P: Progress towards representation of performance data for HVAC&R equipment, in: IBPSA, 2021: pp. 1349–1356. https://doi.org/10.26868/25222708.2021.30280.

[19] Representation of Performance Data for HVAC&R and Other Facility Equipment, (2022).

[20] ASHRAE Standard 232: Common Content and Specifications for Building Data Schemas, (2024).

[21] gbXML, GbXML, (n.d.). https://www.gbxml.org (accessed January 1, 2026).

[22] P. Pauwels, G. Fierro, A Reference Architecture for Data-Driven Smart Buildings Using Brick and LBD

Ontologies, CLIMA 2022 Conf. (2022). https://doi.org/10.34641/clima.2022.425.
[23] M. Pritoni, D. Paine, G. Fierro, C. Mosiman, M. Poplawski, A. Saha, J. Bender, J. Granderson, Metadata Schemas and Ontologies for Building Energy Applications: A Critical Review and Use Case Analysis, Energies 14 (2021) 2024. https://doi.org/10.3390/en14072024.
[24] ISO 16739-1:2018—Industry Foundation Classes (IFC) for Data Sharing in the Construction and Facility Management Industries—Part 1: Data Schema, (2018).
[25] B. Balaji, A. Bhattacharya, G. Fierro, J. Gao, J. Gluck, D. Hong, A. Johansen, J. Koh, J. Ploennigs, Y. Agarwal, M. Bergés, D. Culler, R.K. Gupta, M.B. Kjærgaard, M. Srivastava, K. Whitehouse, Brick : Metadata schema for portable smart building applications, Appl. Energy 226 (2018) 1273–1292. https://doi.org/10.1016/j.apenergy.2018.02.091.
[26] N. Luo, G. Fierro, Y. Liu, B. Dong, T. Hong, Extending the Brick schema to represent metadata of occupants, Autom. Constr. 139 (2022) 104307. https://doi.org/10.1016/j.autcon.2022.104307.
[27] J. Ploennigs, A. Ba, P. Palmes, Extending Brick for automated comfort diagnosis, Automatisierungstechnik 65 (2017) 620–629.
[28] K. Chia, M. Ben-Ayed, S.H. Eshwar, C. Zhang, E. Chen, Y. Gu, J. Kleissl, A. Khurram, J. Wolf, A. Van Sant, K. Trenbath, Brick Schema Standardized Plug Load Control Strategies for Load Reduction: Preprint, National Renewable Energy Laboratory (NREL), Golden, CO (United States), 2024. https://www.osti.gov/biblio/2446843 (accessed February 1, 2026).
[29] X. Jia, Y. Pan, R. Yin, Semantic Modeling and Rule-Based Evaluation of Building Chiller Plant Operation Using Brick Ontology, in: Proc. 12th ACM Int. Conf. Syst. Energy-Effic. Build. Cities Transp., Association for Computing Machinery, New York, NY, USA, 2025: pp. 431–434. https://doi.org/10.1145/3736425.3772320.
[30] J.J. “Jack” M. Gowan, Project Haystack Data Standards, in: Energy Anal., River Publishers, 2015.
[31] H. Li, T. Hong, A semantic ontology for representing and quantifying energy flexibility of buildings, Adv. Appl. Energy 8 (2022) 100113. https://doi.org/10.1016/j.adapen.2022.100113.
[32] G.F. Schneider, P. Pauwels, S. Steiger, Ontology-Based Modeling of Control Logic in Building Automation Systems, IEEE Trans. Ind. Inform. 13 (2017) 3350–3360. https://doi.org/10.1109/TII.2017.2743221.
[33] H. Qiu, G.F. Schneider, T. Kauppinen, S. Rudolph, S. Steiger, Reasoning on human experiences of indoor environments using semantic web technologies, in: Proc. 35th Int. Symp. Autom. Robot. Constr. ISARC 2018 Berl. Ger. July 20-25 2018, International Association on Automation and Robotics in Construction (IAARC), 2018: pp. 95–102. https://doi.org/10.22260/ISARC2018/0013.
[34] M.H. Rasmussen, M. Lefrançois, G.F. Schneider, P. Pauwels, BOT: The building topology ontology of the W3C linked building data group, Semantic Web 12 (2020) 143–161. https://doi.org/10.3233/SW-200385.
[35] N.M. Tomašević, M.Č. Batić, L.M. Blanes, M.M. Keane, S. Vraneš, Ontology-based facility data model for energy management, Adv. Eng. Inform. 29 (2015) 971–984. https://doi.org/10.1016/j.aei.2015.09.003.
[36] I. Fernández del Amo, J.A. Erkoyuncu, D. Bułka, M. Farsi, D. Ariansyah, S. Khan, S. Wilding, Advancing Fault Diagnosis Through Ontology-Based Knowledge Capture and Application, IEEE Access 12 (2024) 144599–144620. https://doi.org/10.1109/ACCESS.2024.3433412.
[37] Y. Chen, G. Lin, E. Crowe, J. Granderson, Development of a Unified Taxonomy for HVAC System Faults, Energies 14 (2021) 5581. https://doi.org/10.3390/en14175581.
[38] Y. Chen, E. Crowe, G. Lin, J. Granderson, What’s in a Name? Developing a Standardized Taxonomy for HVAC System Faults, (2022). https://escholarship.org/uc/item/351568bv (accessed March 1, 2022).
[39] M.Y. Hwang, B. Akinci, M. Berges, FSBrick: An information model for representing fault-symptom relationships in HVAC systems, in: Proc. 10th ACM Int. Conf. Syst. Energy-Effic. Build. Cities Transp., Association for Computing Machinery, New York, NY, USA, 2023: pp. 69–78. https://doi.org/10.1145/3600100.3623729.
[40] M.Y. Hwang, B. Akinci, M. Bergés, FSBrick: an information model for representing fault-symptom relationships in heating, ventilation, and air conditioning systems, Data-Centric Eng. 5 (2024) e33. https://doi.org/10.1017/dce.2024.26.
[41] S. Blechmann, H. Görigk, R. Streblow, D. Müller, Ontology-based approach for fault detection and diagnosis and fault location assessment in air handling units, in: 2024.
[42] A. Hosseini Gourabpasi, M. Nik-Bakht, An ontology for automated fault detection & diagnostics of HVAC using BIM and machine learning concepts, Sci. Technol. Built Environ. 30 (2024) 972–988. https://doi.org/10.1080/23744731.2024.2363104.
[43] J. Ploennigs, M. Maghella, A. Schumann, B. Chen, Semantic Diagnosis Approach for Buildings, IEEE Trans. Ind. Inform. 13 (2017) 3399–3410. https://doi.org/10.1109/TII.2017.2726001.

[44] A. Mallak, A. Behravan, C. Weber, M. Fathi, R. Obermaisser, A Graph-Based Sensor Fault Detection and Diagnosis for Demand-Controlled Ventilation Systems Extracted from a Semantic Ontology, in: 2018 IEEE 22nd Int. Conf. Intell. Eng. Syst. INES, 2018: pp. 000377–000382. https://doi.org/10.1109/INES.2018.8523895.
[45] T. Li, Y. Zhao, C. Zhang, K. Zhou, X. Zhang, A semantic model-based fault detection approach for building energy systems, Build. Environ. 207 (2022) 108548. https://doi.org/10.1016/j.buildenv.2021.108548.
[46] M. Cristani, R. Cuel, A Survey on Ontology Creation Methodologies, Int. J. Semantic Web Inf. Syst. IJSWIS 1 (2005) 49–69. https://doi.org/10.4018/jswis.2005040103.
[47] E.F. Aminu, I.O. Oyefolahan, M.B. Abdullahi, M.T. Salaudeen, A Review on Ontology Development Methodologies for Developing Ontological Knowledge Representation Systems for various Domains, (2020). https://doi.org/10.5815/ijieeb.2020.02.05.
[48] Y. Alfaifi, Ontology Development Methodology: A Systematic Review and Case Study, in: 2022 2nd Int. Conf. Comput. Inf. Technol. ICCIT, 2022: pp. 446–450. https://doi.org/10.1109/ICCIT52419.2022.9711664.
[49] M. Gruninger, S.F. Mark, Methodology for the design and evaluation of ontologies, Proc IJCAI95 Workshop Basic Ontol. Issues Knowl. Shar. (1995).
[50] Y. Li, Z. O'Neill, A critical review of fault modeling of HVAC systems in buildings, Build. Simul. 11 (2018) 953–975. https://doi.org/10.1007/s12273-018-0458-4.
[51] A. Casillas, Y. Chen, J. Granderson, G. Lin, Z. Chen, J. Wen, S. Huang, Development of high-fidelity air handling unit fault models for FDD innovation: lessons learned and recommendations, J. Build. Perform. Simul. 0 (n.d.) 1–16. https://doi.org/10.1080/19401493.2024.2382757.
[52] Y. Hu, D.P. Yuill, S.A. Rooholghodos, A. Ebrahimifakhar, Y. Chen, Impacts of simultaneous operating faults on cooling performance of a high efficiency residential heat pump, Energy Build. 242 (2021) 110975. https://doi.org/10.1016/j.enbuild.2021.110975.
[53] R. Isermann, Model-based fault-detection and diagnosis – status and applications, Annu. Rev. Control 29 (2005) 71–85. https://doi.org/10.1016/j.arcontrol.2004.12.002.
[54] Y. Chen, Data-Driven Whole Building Fault Detection and Diagnosis, Ph.D., Drexel University, 2019. https://www.proquest.com/docview/2275119448/abstract/A8D5FCEB15DC48CAPQ/1 (accessed August 16, 2023).
[55] C. Lu, Z. Wang, M. Mosteiro-Romero, L. Itard, Diagnostic Bayesian network in building energy systems: Current insights, practical challenges, and future trends, Energy Build. 341 (2025) 115845. https://doi.org/10.1016/j.enbuild.2025.115845.
[56] A.A. Amin, K.M. Hasan, A review of Fault Tolerant Control Systems: Advancements and applications, Measurement 143 (2019) 58–68. https://doi.org/10.1016/j.measurement.2019.04.083.
[57] R. Isermann, Fault-Diagnosis Systems: An Introduction from Fault Detection to Fault Tolerance, Springer Science & Business Media, 2005.
[58] G. Lin, M. Pritoni, Y. Chen, J. Granderson, Development and Implementation of Fault-Correction Algorithms in Fault Detection and Diagnostics Tools, Energies 13 (2020) 2598. https://doi.org/10.3390/en13102598.
[59] Y. Zhang, J. Jiang, Bibliographical review on reconfigurable fault-tolerant control systems, Annu. Rev. Control 32 (2008) 229–252. https://doi.org/10.1016/j.arcontrol.2008.03.008.
[60] S.C. Bengea, P. Li, S. Sarkar, S. Vichik, V. Adetola, K. Kang, T. Lovett, F. Leonardi, A.D. Kelman, Fault-tolerant optimal control of a building HVAC system, Sci. Technol. Built Environ. 21 (2015) 734–751. https://doi.org/10.1080/23744731.2015.1057085.
[61] W.G. Jr, HVAC Controls: Operation and Maintenance, CRC Press, 2001.
[62] N. Es-sakali, M. Cherkaoui, M.O. Mghazli, Z. Naimi, Review of predictive maintenance algorithms applied to HVAC systems, Energy Rep. 8 (2022) 1003–1012. https://doi.org/10.1016/j.egyr.2022.07.130.
[63] J. Kim, K. Trenbath, J. Granderson, Y. Chen, E. Crowe, H. Reeve, S. Newman, P. Ehrlich, Research challenges and directions in HVAC fault prevalence, Sci. Technol. Built Environ. 27 (2021) 624–640. https://doi.org/10.1080/23744731.2021.1898243.
[64] N.F. Noy, C.D. Hafner, The State of the Art in Ontology Design: A Survey and Comparative Review, AI Mag. 18 (1997) 53–53. https://doi.org/10.1609/aimag.v18i3.1306.
[65] C. Bezerra, F. Freitas, F. Santana, Evaluating Ontologies with Competency Questions, in: 2013 IEEEWICACM Int. Jt. Conf. Web Intell. WI Intell. Agent Technol. IAT, 2013: pp. 284–285. https://doi.org/10.1109/WI-IAT.2013.199.
[66] S. Staab, R. Studer, Handbook on Ontologies, Springer Science & Business Media, 2013.
[67] Y. Chen, Z. Chen, G. Lin, Y. Zhang, S. Ye, A novel evaluation method of measurement sensitivities on

common faults in VAV HVAC systems, Build. Environ. 261 (2024) 111683. https://doi.org/10.1016/j.buildenv.2024.111683.
[68] J. Granderson, G. Lin, Y. Chen, A. Casillas, J. Wen, Z. Chen, P. Im, S. Huang, J. Ling, A labeled dataset for building HVAC systems operating in faulted and fault-free states, Sci. Data 10 (2023) 342. https://doi.org/10.1038/s41597-023-02197-w.
[69] N. Ghalamsiah, J. Wen, G. Li, Y. Chen, X. Lu, Y. Fu, M. Chu, Z. O'Neill, Labeled Datasets for Air Handling Units Operating in Faulted and Fault-free States, Sci. Data 13 (2026) 15. https://doi.org/10.1038/s41597-025-06179-y.
[70] G. Lin, J. House, Y. Chen, J. Granderson, W. Zhang, Active multi-mode data analysis to improve fault diagnosis in AHUs, Energy Build. 337 (2025) 115621. https://doi.org/10.1016/j.enbuild.2025.115621.
[71] Y. Chen, J. Wen, L.J. Lo, Using Weather and Schedule based Pattern Matching and Feature based PCA for Whole Building Fault Detection — Part II Field Evaluation, ASME J. Eng. Sustain. Build. Cities (2021) 1–16. https://doi.org/10.1115/1.4052730.
[72] J. Huang, H. Yoon, O. Pradhan, T. Wu, J. Wen, Z. O'neill, K.S. Candan, A cosine-based correlation information entropy approach for building automatic fault detection baseline construction, Sci. Technol. Built Environ. 28 (2022) 1138–1149. https://doi.org/10.1080/23744731.2022.2080110.
[73] J. Huang, N. Ghalamsiah, A. Patharkar, O. Pradhan, M. Chu, T. Wu, J. Wen, Z. O'Neill, K. Selcuk Candan, An entropy-based causality framework for cross-level faults diagnosis and isolation in building HVAC systems, Energy Build. 317 (2024) 114378. https://doi.org/10.1016/j.enbuild.2024.114378.
[74] J. Huang, H. Yoon, T. Wu, K.S. Candan, O. Pradhan, J. Wen, Z. O'Neill, Eigen-Entropy: A metric for multivariate sampling decisions, Inf. Sci. 619 (2023) 84–97. https://doi.org/10.1016/j.ins.2022.11.023.
[75] Y. Chen, E. Crowe, J. Granderson, Development of Self-correction Algorithms for Thermostats Using OpenAPI Capabilities, in: Proceeding Int. High Perform. Build. Conf. Purdue, Purdue University, West Lafayette, IN, 2024.
[76] G. Lin, M. Pritoni, Y. Chen, J. Granderson, Development and Implementation of Fault-Correction Algorithms in Fault Detection and Diagnostics Tools, Energies 13 (2020) 2598. https://doi.org/10.3390/en13102598.
[77] S. Wan, M. Zhao, Y. Chen, S. Yang, D. Qiu, L.J. Lo, A novel data-driven relationship inference approach for automatic data tagging in building heating, ventilation and air conditioning systems, Build. Environ. 246 (2023) 110968. https://doi.org/10.1016/j.buildenv.2023.110968.
[78] M. Pritoni, G. Lin, Y. Chen, J. House, E. Crowe, J. Granderson, Market Barriers and Drivers for the Next Generation Fault Detection and Diagnostic Tools, Lawrence Berkeley National Lab. (LBNL), Berkeley, CA (United States), 2022. https://doi.org/10.20357/B7801T.
[79] A. Zeller, Isolating cause-effect chains from computer programs, ACM SIGSOFT Softw. Eng. Notes 27 (2002) 1–10. https://doi.org/10.1145/605466.605468.
[80] C. Lu, C. Struck, C. Miller, D. Saelens, L. Itard, Artificial Intelligence for HVAC Diagnostics: Towards the Era of Large Language Models, EHVA Eur. HVAC J. 2026(2) (n.d.) 59–62.
[81] Y. Chen, J. Wen, A whole building fault detection using weather based pattern matching and feature based PCA method, in: 2017 IEEE Int. Conf. Big Data Big Data, 2017: pp. 4050–4057. https://doi.org/10.1109/BigData.2017.8258421.
[82] Y. Chen, G. Lin, Z. Chen, J. Wen, J. Granderson, A simulation-based evaluation of fan coil unit fault effects, Energy Build. 263 (2022) 112041. https://doi.org/10.1016/j.enbuild.2022.112041.
[83] R. Isermann, Model-based fault-detection and diagnosis – status and applications, Annu. Rev. Control 29 (2005) 71–85. https://doi.org/10.1016/j.arcontrol.2004.12.002.
[84] Y. Chen, J. Wen, T. Chen, O. Pradhan, Bayesian Networks for Whole Building Level Fault Diagnosis and Isolation, in: Int. High Perform. Build. Conf., West Lafayette, IN, 2018: pp. 1–10. https://docs.lib.purdue.edu/ihpbc/266.
[85] B. McBride, The Resource Description Framework (RDF) and its Vocabulary Description Language RDFS, in: S. Staab, R. Studer (Eds.), Handb. Ontol., Springer, Berlin, Heidelberg, 2004: pp. 51–65. https://doi.org/10.1007/978-3-540-24750-0_3.
[86] G. Antoniou, F. van Harmelen, Web Ontology Language: OWL, in: S. Staab, R. Studer (Eds.), Handb. Ontol., Springer, Berlin, Heidelberg, 2009: pp. 91–110. https://doi.org/10.1007/978-3-540-92673-3_4.
[87] H. Li, D. Yu, J.E. Braun, A review of virtual sensing technology and application in building systems, HVACR Res. 17 (2011) 619–645. https://doi.org/10.1080/10789669.2011.573051.
[88] Z. Zhao, X. Han, S. Yang, B. Yang, HVAC Equipment Monitoring, in: Comput. Vis. AI IndoorOutdoor Therm. Environ. Constr. Saf. HVAC Equip. Monit., CRC Press, 2026.

[89] J. Gao, M. Bergés, A large-scale evaluation of automated metadata inference approaches on sensors from air handling units, Adv. Eng. Inform. 37 (2018) 14–30. https://doi.org/10.1016/j.aei.2018.04.010.
[90] N.F. Noy, D.L. McGuinness, Ontology Development 101: A Guide to Creating Your First Ontology, (2001). https://corais.org/sites/default/files/ontology_development_101_aguide_to_creating_your_first_ontology.pdf (accessed May 1, 2025).
[91] K. Jensen, N. Wirth, Pascal User Manual and Report: ISO Pascal Standard, Springer Science & Business Media, 2012.
[92] A. Gómez-Pérez, Ontology Evaluation, in: S. Staab, R. Studer (Eds.), Handb. Ontol., Springer, Berlin, Heidelberg, 2004: pp. 251–273. https://doi.org/10.1007/978-3-540-24750-0_13.
[93] J. Brank, M. Grobelnik, D. Mladenić, A survey of ontology evaluation technique, in: Proc. Conf. Data Min. Data Wareh. SiKDD, Ljubljana, Slovenia, 2005.
[94] J. Wen, S. Li, ASHRAE 1312-RP: Tools for Evaluating Fault Detection and Diagnostic Methods for Air-Handling Units—Final Report, Drexel University, Philadelphia, PA, 2011.
[95] J. Granderson, G. Lin, Y. Chen, A. Casillas, LBNL Fault Detection and Diagnostics Datasets, DOE Open Energy Data Initiative (OEDI); Lawrence Berkeley National Lab. (LBNL), Berkeley, CA (United States), 2022. https://doi.org/10.25984/1881324.
[96] J. Granderson, G. Lin, A. Harding, P. Im, Y. Chen, Building fault detection data to aid diagnostic algorithm creation and performance testing, Sci. Data 7 (2020) 65. https://doi.org/10.1038/s41597-020-0398-6.
[97] S. Jung, Y. Yoon, P. Im, Datasets of Faults in Variable Air Volume Terminal Units in a Multi-Zone Commercial Building, Sci. Data 12 (2025) 763. https://doi.org/10.1038/s41597-025-05063-z.
[98] B. DuCharme, Learning SPARQL: Querying and Updating with SPARQL 1.1, O'Reilly Media, Inc., 2013.

**Nomenclature**

***Abbreviation***

| | |
|---|---|
| FDD | Fault detection and diagnostics |
| FDD-ON | Fault detection and diagnostics-Ontology |
| CQ | Competency question |
| VAV | Variable air volume |
| HVAC | Heating, ventilation, and air conditioning |
| HVAC&R | Heating, ventilation, air conditioning and refrigeration |
| BAS | Building automation system |
| CMMS | Computerized maintenance management system |
| O&M | Operation and maintenance |
| BIM | Building information model |
| AI | Artificial intelligence |
| LLM | Large language model |
| NLP | Natural language process |
| AHU | Air handling unit |
| ATU | Air terminal unit |
| RTU | Packaged rooftop unit |
| OWL | Web Ontology Language |
| W3C | World Wide Web Consortium |
| RDF | Resource Description Framework |
| RDFS | Resource Description Framework Schema |
| DL | Description logics |
| SPARQL | Semantic Protocol and RDF Query Language |
| BN | Bayesian Network |

**Appendix I**

**Table** I.1 Feature comparison between FDD-ON and other ontologies or semantic models

| **Feature** | **IFC [24]** | **Project Haystack [22]** | **Brick Schema [17]** | **FSBrick [29, 30]** | **AFDDOnto [32]** | **Fault taxonomy [13]** | **FDD-ON** |
|---|---|---|---|---|---|---|---|
| **Model** | | | | | | | |
| **--Modeling approach** | **Tag-based** | **Tag-based** | **Ontology-based** | **Ontology-based** | **Ontology-based** | **Tag-based** | **Ontology-based** |
| **--Formal semantics** | **Y** | **Y** | **Y** | **Y** | **Y** | **Y** | **Y** |
| **--Query language** | Haxall / filters | Haxall / filters | SPARQL | SPARQL | SPARQL | Python/Filters | SPARQL |
| **--Controlled vocabulary** | | **Y** | **Y** | | | **Y** | **Y** |
| **Asset** | | | | | | | |
| **--Fault asset** | | | | | | | |
| -----Fault library | | | | | | **Y** | **Y** |
| -----Description of a fault type | | | | | | | **Y** |
| -----Information specifying the physical location at which a fault occurs | | | | | **Y** | **Y** | **Y** |
| -----Information specifying the component type where the fault occurs | | | | **Y** | | **Y** | **Y** |
| -----Information specifying the fault mode | | | | **Y** | | **Y** | **Y** |
| -----Information quantifying the extent of fault severity level | | | | | | | **Y** |
| -----Information specifying fault contributing causes | | | | | | | **Y** |
| **--Symptom asset** | | | | | | | |
| -----Symptom library | | | | | | **Y** | **Y** |
| -----Description of a symptom type | | | | | | | **Y** |
| -----Information specifying the location of the observed symptom | | | | **Y** | | | **Y** |
| -----Information specifying measures used to generate symptoms | | **Y** | **Y** | **Y** | | | **Y** |
| -----Information specifying a symptom status | | | | **Y** | | **Y** | **Y** |
| **--Impact asset** | | | | | | | |
| -----Impact library | | | | | | | **Y** |
| **Taxonomy** | | | | | | | |
| **--Hierarchical classification** | | **Y** | **Y** | | | **Y** | **Y** |
| **--Data structure** | | | | | | | |
| -----Structured data to represent a fault | | | | **Y** | | **Y** | **Y** |
| -----Structured data to represent a symptom | | | | | | | **Y** |
| **--Fault attribute** | | | | | | | |
| -----Classification of fault behaviors | | | | Limited | | | **Y** |

| -----Classification of component types | | | | | | | **Y** |
|---|---|---|---|---|---|---|---|
| -----Classification of fault contributing causes | | | | | | | ***Y*** |
| -----Classification of fault lifecycle occurrence stages | | | | | | | **Y** |
| **--Symptom attribute** | | | | | | | |
| -----Classification of symptom data types | | | | | | | **Y** |
| -----Classification of symptom patterns | | | | | | | **Y** |
| **--Impact attribute** | | | | | | | |
| -----Classification of impact types | | | | | | | **Y** |
| **Data relations** | | | | | | | |
| --Fault and symptom relation | | | | **Y** | | **Y** | **Y** |
| --Fault and impact relation | | | | | | **Y** | **Y** |
| --Fault and corresponding attributes | | | | | | | **Y** |
| --Symptom and corresponding attributes | | | | | | | **Y** |